%% file: main.tex
\documentclass[runningheads]{llncs}

\usepackage[mobile]{eccv}

\usepackage{eccvabbrv}
\usepackage{graphicx}
\usepackage[accsupp]{axessibility}  
\usepackage{capt-of,xcolor,xspace,enumitem}
\usepackage{amsmath,amssymb,amsbsy,amsfonts,dsfont,pifont,bm,bbm,mathrsfs,mathtools,nicefrac,marvosym}
\usepackage{algorithm,algpseudocode,listings}
\usepackage{booktabs,multirow,adjustbox,diagbox,threeparttable,tabularray}

\usepackage[pagebackref,breaklinks,colorlinks,citecolor=eccvblue]{hyperref}
\usepackage{hyperref}  

\usepackage{orcidlink}
\usepackage{wrapfig}
\usepackage[capitalize]{cleveref}  
\crefname{section}{Sec.}{Secs.}
\Crefname{section}{Section}{Sections}
\crefname{table}{Tab.}{Tabs.}
\Crefname{table}{Table}{Tables}
\crefname{figure}{Fig.}{Figs.}
\Crefname{figure}{Figure}{Figures}
\crefname{equation}{Eq.}{Eqs.}
\Crefname{equation}{Equation}{Equations}
\newcommand{\tocite}[1]{{\color{red} [TO CITE]}}
\newcommand{\method}{{\texttt{FlashFace}}\xspace}  
\newcommand{\supp}{\textit{Supplementary Material}\xspace}

\newcommand{\shilong}[1]{}

\newcolumntype{x}[1]{>{\centering\arraybackslash}p{#1}}
\newcolumntype{y}[1]{>{\raggedright\arraybackslash}p{#1}}
\newcolumntype{z}[1]{>{\raggedleft\arraybackslash}p{#1}}

\begin{document}

\title{\includegraphics[scale=0.5]{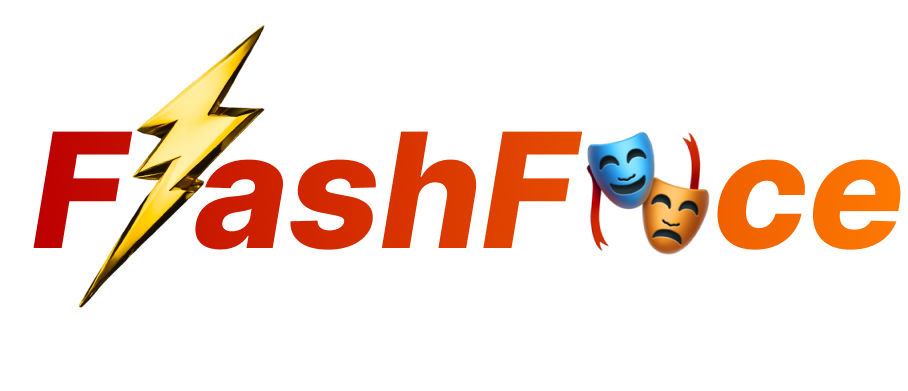} \\ 
\vspace{-6mm}
Human Image Personalization with High-fidelity Identity Preservation
\vspace{-5mm}
}

\titlerunning{\method}

\author{
 \small    Shilong Zhang\inst{1}\and
    Lianghua Huang\inst{2} \and
    Xi Chen\inst{1} \and
    Yifei Zhang \inst{2}  \\
 \small Zhi-Fan Wu \inst{2} \and 
    Yutong Feng \inst{2} \and
    Wei Wang \inst{2} \and
    Yujun Shen  \inst{3} \and
    Yu Liu \inst{2} \and
    Ping Luo \inst{1}
}
\authorrunning{Zhang et al.}

\institute{
  \small  $^{1}$The University of Hong Kong, $^{2}$Alibaba Group, $^{3}$Ant Group
}

\maketitle

\vspace{-5mm}

\input{sections/0.abs}
\input{sections/1.intro}

\input{sections/2.relatated_work}

\input{sections/3.method}

\input{sections/4.exp}
\input{sections/5.conclusion}
\input{sections/6.ref}

\title{\method: Human Image Personalization with High-fidelity Identity Preservation\\ (\supp)}

\titlerunning{\method}

\author{
 \small    Shilong Zhang\inst{1}\and
    Lianghua Huang\inst{2} \and
    Xi Chen\inst{1} \and
    Yifei Zhang \inst{2}  \\
 \small Zhi-Fan Wu \inst{2} \and 
    Yutong Feng \inst{2} \and
    Wei Wang \inst{2} \and
    Yujun Shen  \inst{3} \and
    Yu Liu \inst{2} \and
    Ping Luo \inst{1}
}
\authorrunning{Zhang et al.}

\institute{
  \small  $^{1}$The University of Hong Kong, $^{2}$Alibaba Group, $^{3}$Ant Group
}

\maketitle

\input{sections/7.appendix}

\end{document}

%% file: sections/0.abs.tex
\begin{abstract}

This work presents \method, a practical tool with which users can easily personalize their own photos on the fly by providing one or a few reference face images and a text prompt.
Our approach is distinguishable from existing human photo customization methods by \textit{higher-fidelity identity preservation} and \textit{better instruction following}, benefiting from two subtle designs.
First, we encode the face identity into a series of feature maps instead of one image token as in prior arts, allowing the model to retain more details of the reference faces (\textit{e.g.}, scars, tattoos, and face shape ).
Second, we introduce a disentangled integration strategy to balance the text and image guidance during the text-to-image generation process, alleviating the conflict between the reference faces and the text prompts (\textit{e.g.}, personalizing an adult into a ``child'' or an ``elder'').
Extensive experimental results demonstrate the effectiveness of our method on various applications, including human image personalization, face swapping under language prompts,  making virtual characters into real people, \textit{etc.} 
The page of this project is ~\href{https://jshilong.github.io/flashface-page}{https://jshilong.github.io/flashface-page}.


\end{abstract}

%% file: sections/1.intro.tex
\begin{figure}[t]
    \includegraphics[width=1.0\linewidth]{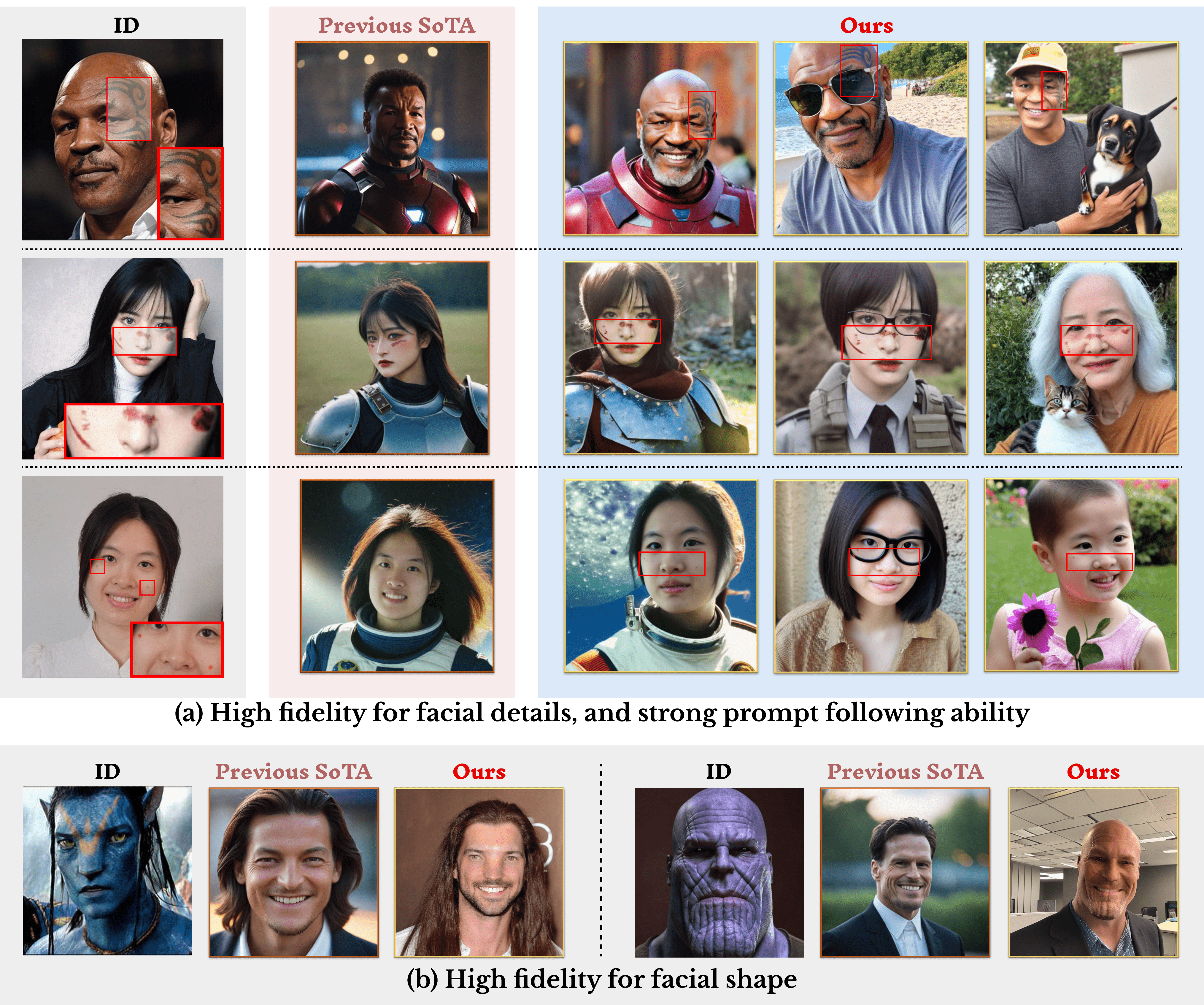}
    \vspace{-18pt}
    \caption{%
        \textbf{Diverse human image personalization results} produced by our proposed \method, which enjoys the features of (1) preserving the identity of reference faces \textit{in great details} (\textit{e.g.}, tattoos, scars, or even the rare face shape of virtual characters) and (2) \textit{accurately following the instructions} especially when the text prompts contradict the reference images (\textit{e.g.}, customizing an adult to a ``child'' or an ``elder''). Previous SoTA refers to PhotoMaker~\cite{li2023photomaker}.
    }
    \label{fig:teaser}
    \vspace{-10pt}
\end{figure}

\vspace{-5mm}
\section{Introduction}

Image generation has advanced significantly in recent years. The overall framework has evolved from GANs~\cite{goodfellow2014generative, radford2016unsupervised} to Diffusion models~\cite{ho2020denoising, dhariwal2021diffusion}, with improved training stability and generation quality. At the same time, large-scale multi-modal datasets~\cite{schuhmann2022laion5b, radford2021learning} containing image-caption pairs connect the tasks of visual and linguistic processing. In this way, text starts to act as a control signal for image generation, bringing algorithms and products such as Stable Diffusion~\cite{rombach2021highresolution, podell2023sdxl}, DALL-E series to the forefront and sparking a massive productivity revolution with diversified applications.

\begin{figure}[t]
    \centering
    \includegraphics[width=0.8\linewidth]{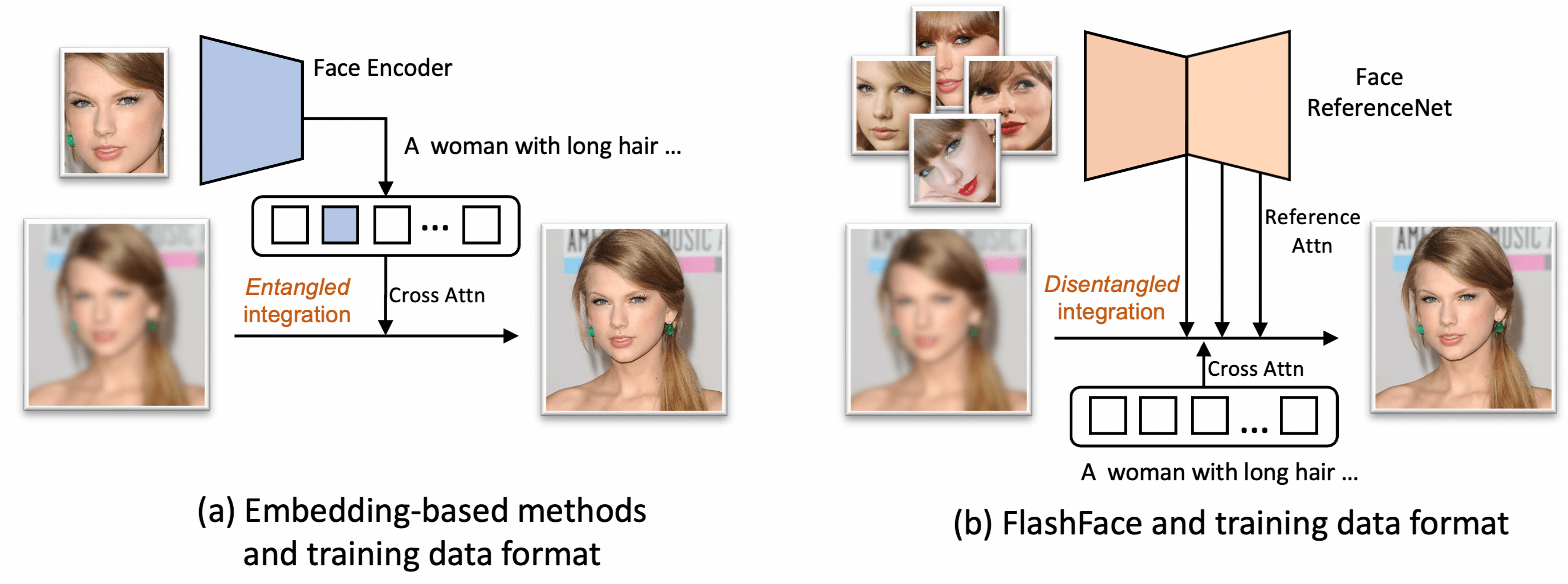}
    \vspace{-8pt}
    \caption{
        \textbf{Concept comparison} between \method and previous embedding-based methods.  We encode the face to a series of feature maps instead of several tokens to preserve finer details. We do the disentangled integration using separate layers for the reference and text control, which can help to achieve better instruction following ability. We also propose a novel data construction pipeline that ensures facial variation between the reference face and the generated face.
    }
    \label{fig:teaser2}
    \vspace{-8pt}
\end{figure}

Among the applications of image generation, human image customization is a popular and challenging topic.
It requires generating images with consistent face identity. Rather than relying solely on language conditions, this task integrates the human identity from multiple reference face images to guide the generation. Early explorations~\cite{ruiz2023dreambooth, gal2022image, hu2021lora} require test-time fine-tuning, which takes 
about half an hour to achieve satisfactory results for each user. Recent methods~\cite{xiao2023fastcomposer, ye2023ip-adapter, yan2023facestudio, li2023photomaker} investigate the zero-shot setting to make the model work on the fly(usually seconds). As shown in \cref{fig:teaser2},  these methods extract face ID information using a face encoder and convert it into one or several tokens. These tokens are then injected into the generation process along with text tokens. To gather sufficient training data, they also simplify data collection by cropping the face region from the target image as a reference.

These zero-shot methods make human image customization practical with significantly faster speed.
However, they still have two limitations. 
First, they struggle to preserve the face shape and details.
This is due to the loss of spatial representations when encoding the reference faces into one or a few tokens. Secondly, it is challenging for these methods to achieve precise language control. 
For instance, they usually fail to follow the prompt when the prompt requests the generation of an elderly person but provides a young person as the reference image. 
We analyze that this issue is caused by treating face tokens and textual tokens equally 
and integrating them into U-Net at the same position. This design makes the two types of control signals entangled.
Additionally, their data construction pipeline, which involves cropping the target image to construct a reference face image, leads the model to prioritize copying reference images rather than following the language prompt.

This study addresses these two major limitations at once.  First, as shown in \cref{fig:teaser2}, to enhance facial identity fidelity, we leverage a reference network~\cite{Reference-only} to encode the reference image into a series feature map. 
Compared with the ``token representation'' applied by previous methods~\cite{ye2023ip-adapter,xiao2023fastcomposer,yan2023facestudio,li2023photomaker}, these feature maps retain the spatial shape and encompass richer facial shape and detail information(as shown in ~\cref{fig:teaser}). 
Second, to balance the control signal of the reference image and the text prompt, we inject them in a disentangled manner. 
Specifically, we insert additional reference attention layers in the U-Net decoder to integrate the reference features~(as shown in \cref{fig:teaser2}). 
These layers are separate from the cross-attention layer used for text conditions, ensuring that the two signals are disentangled.  
We also allow users to adjust the weight of the features produced by the reference attention layers.
When incorporating this re-weighing operation with the classifier-free guidance~\cite{ho2022classifier}, we achieve a smooth control of the face reference intensity. 
Besides the improvements of the model structure, we also propose a new data construct pipeline that gathers multiple images of a single person. 
This kind of data ensures the variation between the faces of the reference image and the target image during training, thereby mitigating the network's tendency to directly copy the reference face. In this way, the model is forced to acquire guidance from the text prompt, and the prompt-following ability could be further improved.

In summary, our work makes the following contributions:
\begin{itemize}[leftmargin=*]
    \item  We propose to encode the reference face image into a series of feature maps instead of one or several tokens to maintain more details, which assist the model in generating high-fidelity results.
    \item  We incorporate reference and text control signals in a disentangled manner with separate layers. Together with our new data construction pipeline, we achieve precise language control, even in cases where there is a conflict between the prompt and the reference face images.
  \item We demonstrate the remarkable effectiveness of \method across a wide range of downstream tasks, including human image customization, face swapping under different language prompts, bringing virtual characters as real people, etc. These capabilities open up infinite possibilities for the community.
        
\end{itemize}

%% file: sections/2.relatated_work.tex
\section{Related Work}

\noindent \textbf{Text-to-Image Generation.} Text-to-image generation~\cite{goodfellow2014generative, radford2016unsupervised,stylegan, scalinggan,unclip,imagen,rombach2021highresolution, chen2023gentron} has significantly progressed in recent years, which can be primarily attributed to the development of diffusion models~\cite{sohl2015diffusion,ncsn,ho2020denoising,dhariwal2021diffusion,podell2023sdxl} and auto-regressive models~\cite{ramesh2021zero,muse}.
The state-of-the-art text-to-image generation models are capable of generating images precisely following text prompts provided by the user~\cite{imagen,podell2023sdxl}. Meanwhile, the generated images are remarkably realistic while maintaining a high level of aesthetic appeal~\cite{unclip}. 
However, these existing methods are limited to generating images solely from text prompts, and they do not meet the demand for efficiently producing customized images with the preservation of identities.

\noindent \textbf{Subject-Driven Image Generation.} Subject-driven image generation focuses on creating new images of a particular subject based on user-provided images. Methods like DreamBooth~\cite{ruiz2023dreambooth} and textual-inversion~\cite{gal2022image} along with their subsequent works~\cite{kumari2023multiconcept,wei2023elite,liu2023cones,liu2023cones2} utilize optimization-based approaches to embed subjects into diffusion models. However, these methods usually necessitate approximately 30 minutes of fine-tuning per subject, posing a challenge for widespread utilization. Zero-shot subject-driven image generation~\cite{ye2023ip-adapter, chen2023anydoor, pan2023kosmos, Emu2} aims to generate customized images without extra model training or optimization process. Nevertheless, an obvious issue is that compared to optimization-based methods, these zero-shot approaches often sacrifice the ability to preserve identity, i.e., the generated images could lose the fine details of the original subject.

\noindent \textbf{Human Image Personalization.} In this paper, we mainly focus on the identity preservation of human faces, as it is one of the most widely demanded and challenging directions in subject-driven image generation. Recently, methods like IP-Adapter-FaceID~\cite{ye2023ip-adapter}, FastComposer~\cite{xiao2023fastcomposer}, and PhotoMaker~\cite{li2023photomaker} show promising results on zero-shot human image personalization. 
They encode the reference face to one or several tokens as conditions to generate customized images. Yet, these methods usually fail to maintain the details of human identities. In contrast, our proposed method precisely retains details in human faces and accurately generates images according to text prompts in a zero-shot manner.

%% file: sections/3.method.tex
\section{Method}

In this section, we first introduce a universal pipeline for the construction of person ID datasets. Distinct from previous dataset~\cite{karras2019stylebased, liu2023hyperhuman, zheng2021farl}, we incorporate individual ID annotation for each image, allowing the sampling of multiple images of the same individual throughout the training process. The tools used in this pipeline are publicly accessible on the Internet, simplifying replication for fellow researchers. Next, we present \method framework, which efficiently preserves facial details, and enables high-fidelity human image generation.

\subsection{ID Data Collection Pipeline}

We introduce a generic methodology for assembling a person ID dataset, which includes numerous images of each individual. Our approach involves data collection, data cleaning, and annotation. On average, the dataset presents approximately one hundred images per person, ensuring a diverse range of expressions, accessories, hairstyles, and poses. We believe that this pipeline will greatly contribute to the advancement of human image personalization.

\noindent\textbf{Data collection.} 
Our initial step is to collect public figures who are well-represented online, including actors, musicians, and other celebrities. To construct our ID list, we extract names from ranking platforms like IMDb~\cite{IMDb}, Douban~\cite{douban}, YouGov~\cite{YouGov}, among others. Utilizing these names, we construct a rich repository of publicly accessible images. To capture a wide set of visual data, we collect at least 500 images for each individual.

\noindent\textbf{Data cleaning.}
To remove noise images that feature individuals other than the intended person within each ID group, we employ a face detection model (Retinaface-R50~\cite{Deng_2020_CVPR}) to detect all faces. We then utilize a face recognition model (ArcFace-R101~\cite{Deng_2022}) to obtain face embeddings. By applying the k-means clustering algorithm, we select the
center embedding of the largest cluster as the indicator embedding for the intended person. Subsequently, we filter out similarity scores below a specified threshold (0.6) within the largest cluster and all images in other clusters, effectively removing the noise images.

\noindent \textbf{Image annotation.}
To ensure our image annotations are diverse and rich, we leverage Qwen-VL~\cite{Qwen-VL}, a powerful vision-language model that is capable of accurately describing individuals in the images. We prompt it with a specific query: ``Can you provide a detailed description of the person in the photo, including their physical appearance, hairstyle, facial expression, pose or action, accessories, clothing style, and background environment?'' This allows us to generate captions that offer detailed descriptions of images, thus enabling precise control over the appearance of individuals.

\noindent \textbf{Dataset statistics.}
Finally, leveraging our scalable pipeline, we managed to compile a dataset of 1.8 million images featuring 23,042 individuals. Each individual is represented by an average of 80 images, 
providing a rich individual context that enables our model to learn identification features from the reference images instead of simply copying the original image onto target images.
We construct a validation dataset consisting of 200 individuals, while the remaining data is used for training. For additional statistical details about the dataset, please refer to our \supp.

\subsection{\method}

\begin{figure}[t]
    \centering
    \includegraphics[width=0.9\linewidth]{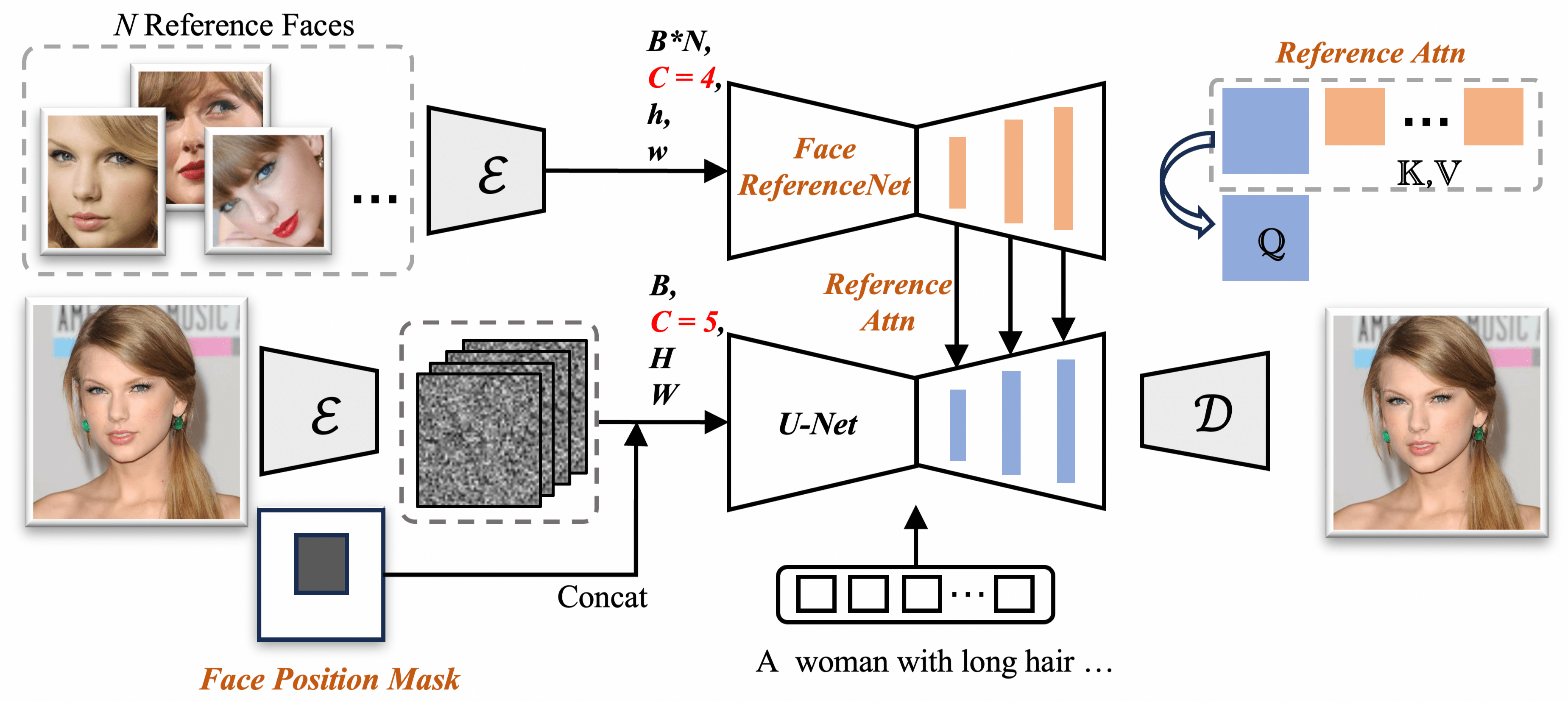}
    \vspace{-7pt}
    \caption{%
        \textbf{The overall pipeline of \method.} During training, we randomly select $B$ ID clusters and choose $N+1$ images from each cluster. We crop the face region from $N$ images as references and leave one as the target image. This target image is used to calculate the loss. The input latent of Face ReferenceNet has shape $(B*N) \times 4 \times h \times w$. We store the reference face features after the self-attention layer within the middle blocks and decoder blocks. A face position mask is concatenated to the target latent to indicate the position of the generated face. During the forwarding of the target latent through the corresponding position in the U-Net, we incorporate the reference feature using an additional reference attention layer. During inference, users can obtain the desired image by providing a face position(optional), reference images of the person, and a description of the desired image.
    }
    \label{fig:arch}
    \vspace{-10pt}
\end{figure}

As shown in \cref{fig:arch}, the overall framework is based on the widely used SD-V1.5 project~\cite{Rombach_2022_CVPR}. All images are projected into the latent space with a stride 8 auto-encoder~\cite{kingma2013auto}. A U-Net~\cite{ronneberger2015u} is employed for denoise. The language branch of CLIP~\cite{radford2021learning} is introduced to encode the language prompt and then use the cross-attention operation to do the integration. A notable feature of our framework is the inclusion of a Face ReferenceNet~\cite{Reference-only}, which extracts detailed facial features that retain the spatial shape and incorporate them into the network using additional reference attention layers.

\noindent\textbf{Face ReferenceNet.} We utilize an identical U-Net architecture with the same pre-trained parameters (SD-V1.5~\cite{Rombach_2022_CVPR}) as our Face ReferenceNet. During training,  we randomly sample $B$ ID clusters, and then random sample $N$ reference images from each ID cluster. We crop the face regions and resize them to  $224\times224$. After VAE encoding, these faces are converted to the input latents with the shape of $(B*N)\times 4\times 28\times 28$. As the latent feature passes through the middle and decoder blocks, we cache all features after the self-attention layer, which have shape $(B*N)\times (h_r*w_r)\times C $. The number of reference images $N$  is randomly set between 1 and 4 during training.

\noindent\textbf{U-Net with reference attention layer.}
During the training phase, we encode the selected $B$ target images as initial latent representations with 4 channels. For our face inpainting model, we fill the latent representation inside the face region with 0 and concatenate it with the noised latent representation to form the input with 8 channels. To facilitate the composition for the user, we initialize a face position mask, which is a one-channel zero mask only the region within the face bounding box filled 1. The face mask is then connected to the noised latent representation as the input latent. When the latent feature forwards to the same position as the cached reference features, we use an extra reference attention layer to incorporate reference face features into the U-Net. As shown in \cref{equ:ref}, the query(\textit{Q}) remains as the U-Net latent feature($y_{g}$), which shapes  $B  \times (h_g * w_g) \times C$, while the key(\textit{K}) and value(\textit{V}) are concatenated features consisting of the U-Net and all reference face features($y_{ri}$, $i$ refer to the $i_{th}$ reference face image). All the reference features are contacted with the U-Net feature along the spatial dimension, and get the \textit{K},\textit{V} with length $h_g * w_g + N*h_r*w_r$. To maintain the original image generation capability, we randomly drop the face position mask and reference feature with a probability of 0.1 during training.
\begin{equation} y_{g} = y_{g} + \begin{cases} 
Attn(\textit{Q}=y_{g},\textit{K},\textit{V}=y_{g}), &\quad\text{if $random$} < 0.1 \\ Attn(\textit{Q}=y_{g},\textit{K},\textit{V}=cat[y_{g},y_{r1}, y_{r2} ...]), &\quad\text{if $random$}\geq 0.1 \\ \end{cases} 
\label{equ:ref}
\end{equation}
\noindent \textbf{Face reference strength.} We also provide one way to control the reference strength in the inference phase. This allows users to adjust the balance of the language prompt and references in case of conflicts. For example, to make a person a bronze statue as seen in \cref{fig:ref_strength}.

Firstly, in \cref{equ:strenghth}, we introduce $\lambda_{feat}$ to re-weight the two reference attention layer results: One result incorporates the reference features in its $\textit{K}$ and $\textit{V}$, while the other does not. A higher value of $\lambda_{feat}$ assigns greater weight to the result with reference feature, resulting in a higher ID fidelity face in the generated images. This operation provides smooth control over the reference strength. It also facilitates the \textbf{identity mixing} by easily changing to re-weight attention results associated with two different individuals, we show the case in our \supp.
\begin{equation}
y_{g}  = y_{g} + (1- \lambda_{feat}) *Attn(y_{g}, y_{g}) + \lambda_{feat}*Attn(y_{g}, cat[y_{g}, y_{r1}, y_{r2} ...]) 
\label{equ:strenghth}
\end{equation}
Secondly, we also involve classifier-free guidance~\cite{ho2022classifier} by combining three inference results under different conditions: no reference image or text prompt ($Y_{None}$), only text prompt ($Y_{Text}$), and both text prompt and reference image ($Y_{Text\&Ref}$). The final results are obtained by combining these results using the \cref{equ:guidence}, and $\lambda_{text}$, $\lambda_{ref}$ are weighting factors for text guidance and reference guidance.  Reference guidance is not that smooth like  $\lambda_{feat}$, but it helps to preserve more details(refer to \supp). In this study, $\lambda_{feat}=0.85$,  $\lambda_{ref}=2$, and $\lambda_{text}=7.5$ by default. 
\begin{equation}
Y = Y_{None} +\lambda_{text} * (Y_{Text} - Y_{None}) + \lambda_{ref} * (Y_{Text\&Ref} - Y_{Text}) 
\label{equ:guidence}
\end{equation}

%% file: sections/4.exp.tex
\section{Experiments}

\subsection{Implementation details}

In this study, we employ the SD-V1.5 model~\cite{rombach2021highresolution}. For both human customization and face inpainting models, we use Adam~\cite{kingma2014adam} to do the parameter optimization on 8 NVIDIA A100 GPUs, with a batch size of 32, for a total of 50000 iterations. The learning rate is set to 2e-5 for both the Face ReferenceNet and U-Net modules. During training, we randomly select $N$ reference images and one target image from the same ID cluster for training purposes($N$  is randomly set between 1 and 4). We crop the face region of the reference images and resize them to $224 \times 224$. The target images are also resized to $768 \times 768$ to represent the face details better. To improve generation quality through classifier-free guidance~\cite{ho2022classifier}, there is a 10\% chance of using a null-text embedding. For sampling, we utilize 50 steps of the DDIM sampler~\cite{song2020denoising}. The feature strength $\lambda_{feat}$ is set to 0.85 by default. The scale of text classifier-free guidance~\cite{ho2022classifier}  is set to 7.5 by default, and the reference guidance is set to 2 by default.

\subsection{Applications}
\label{sec.app}
In this section, we demonstrate four applications that highlight the strong customization ability of our method. With our novel human customization model, we can generate realistic human images of different attributes, ages, and genders while keeping the identity of individuals. Additionally, we can do the transformation between the artwork and real people with high face ID fidelity. Finally, we showcase the impressive performance of our face inpainting model, which seamlessly places your face to anybody even with an additional language prompt. \method and PhotoMaker~\cite{li2023photomaker} utilize four reference images as input, while the remaining methods use only one reference image as input. The reference images for each figure can be found in our \supp.

\begin{figure}[t]
    \centering
    \includegraphics[width=\linewidth]{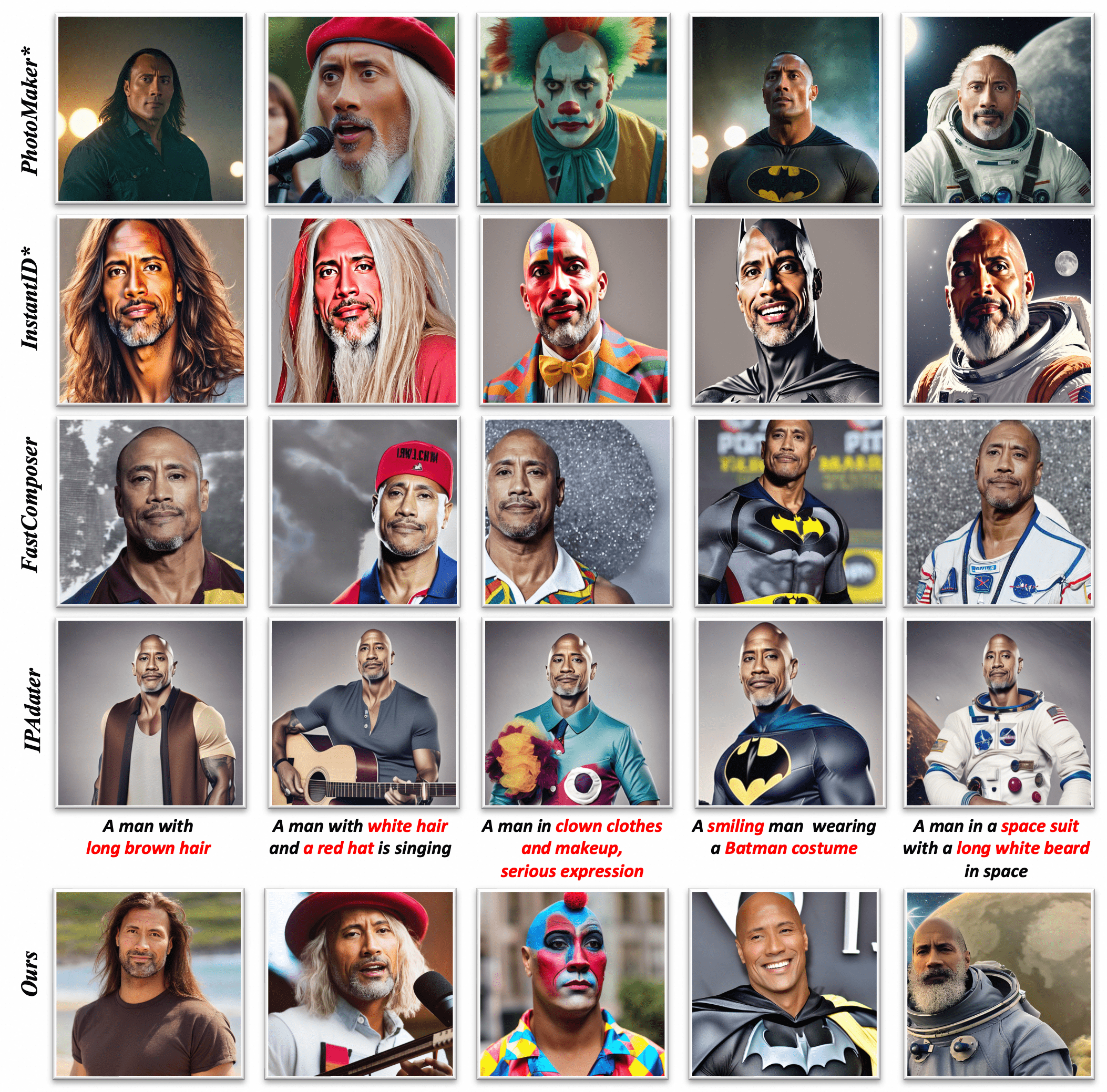}
    \vspace{-10pt}
    \caption{%
        \textbf{Image personalization results of ``Dwayne Johnson''.} \method maintains high fidelity and accurately follows the language prompt. The methods marked with $*$ are our concurrent works.
    }
    \label{fig:general}
    \vspace{-16pt}
\end{figure}

\noindent\textbf{General human image customization.} As shown in \cref{fig:general}, we present image generation results of  ``Dwayne Johnson'' using different language prompts. We note a tendency of copy-pasting and same appearance among the generated faces of IP-Adapter-faceid~\cite{ye2023ip-adapter}, FastComposer~\cite{xiao2023fastcomposer}, and InstantID~\cite{wang2024instantid}. They fail to achieve natural face variations based on the language prompt. For instance, ``singing'' and ``clown makeup''. \method maintains high fidelity and accurately follows the language prompt. With SD-V1.5~\cite{rombach2021highresolution}, and achieve comparable performance to the concurrent work, PhotoMaker~\cite{li2023photomaker}, which uses a much larger model (SDXL~\cite{podell2023sdxl}).

\begin{figure}[t]
    \centering
    \includegraphics[width=1.0\linewidth]{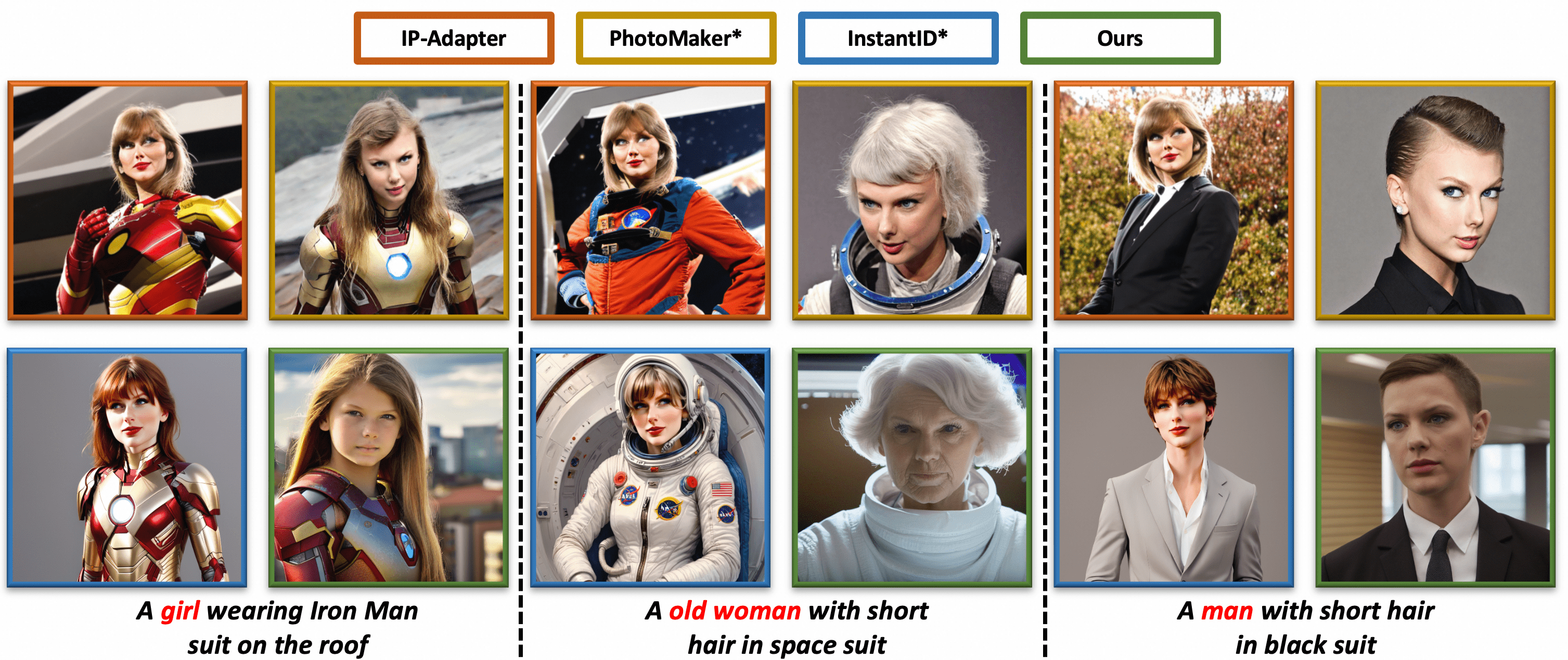}
    \vspace{-18pt}
    \caption{%
        \textbf{Personalization results of ``Taylor Swift'' with different ages and genders}. For different prompts,  the facial details of Photomaker~\cite{li2023photomaker} and InstantID~\cite{wang2024instantid} remain unaltered and continue to resemble a young woman. Our \method seamlessly combines the characteristics represented by age and gender words with the original person ID. The methods marked with $*$ are our concurrent works.
    }
    \label{fig:age_gender}
    \vspace{-10pt}
\end{figure}

\label{sec:attribute}

\noindent\textbf{Change the age and gender.} As shown in \cref{fig:age_gender}, \method demonstrates a remarkable capability to alter age and gender, which conflicts with the reference images. By utilizing four reference images of ``Taylor Swift'' and reference strength $\lambda_{feat}=0.85$ and guidance scale  $\lambda_{ref}=2$, \method can generate images that seamlessly combine the youthful facial features of a ``girl'', the aging wrinkles of ``old'', and the masculine characteristics of a ``man'' into the generated face while maintaining ID information. In comparison, IP-Adapter-faceid~\cite{ye2023ip-adapter} shares the same face part for all three age-related prompts and even fails to maintain facial fidelity. For our concurrent work,  Photomaker~\cite{li2023photomaker} and InstantID~\cite{wang2024instantid} successfully preserve identity information. However, when it comes to different ages and genders, the facial details remain unaltered and continue to resemble a young woman. (Please zoom in to examine the facial details closely)

\begin{figure}[t]
    \centering
    \includegraphics[width=1.0\linewidth]{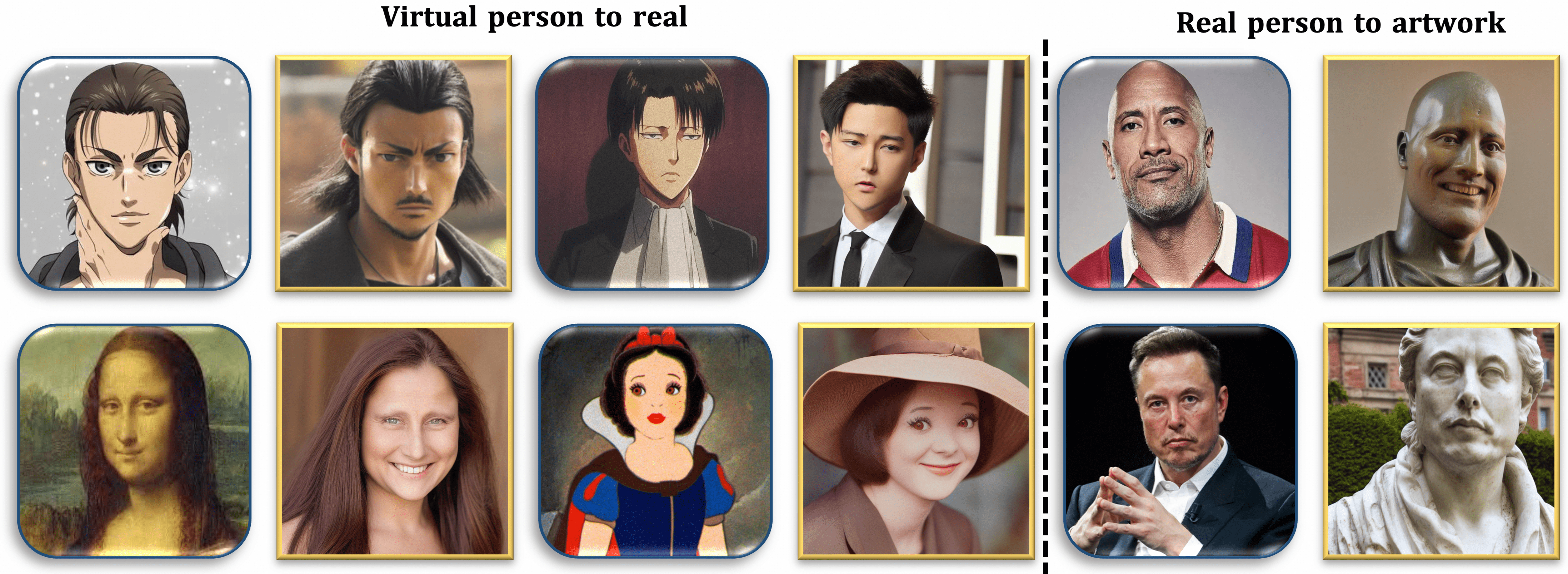}
    \vspace{-18pt}
    \caption{%
       \textbf{Transformation between artwork and real people}. \method even can do transformation between artwork and real people with high face fidelity. We can see the face shape is well preserved in both transformations.
    }
    \label{fig:art}
    \vspace{-10pt}
\end{figure}

\noindent \textbf{Transformation between the artwork and real people.} 
\method also supports the transformation between virtual persons and real individuals with high face-shape fidelity. For virtual persons into real, we only require a single reference image, and a simple description(such as "a man in a black suit"). When converting a real person into an artwork, we utilize four reference images. For simplicity, only one image is displayed in the \cref{fig:art}. Additionally, the prompt should specify the type of artwork, for example, "a marble statue of a man" or "a bronze statue of a man".

\begin{figure}[!h]
    \centering
    \includegraphics[width=1.0\linewidth]{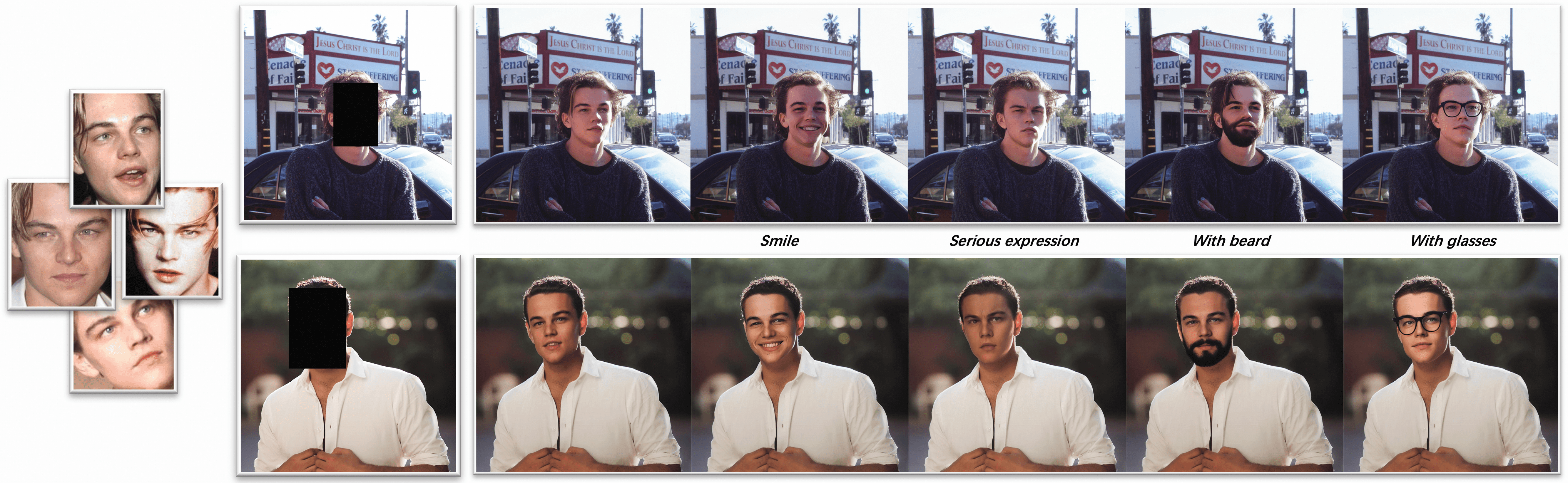}
    \vspace{-18pt}
    \caption{%
       \textbf{ID face inpainting with different prompts.} $\method_{inpainting}$ is capable of inpainting a person's face onto another individual with high face ID fidelity. Additionally, during the inpainting process, it allows using language to specify facial details like facial expressions, glasses, beards, etc.
    }
    \label{fig:inpainting}
    \vspace{-10pt}
\end{figure}

\noindent \textbf{ID face inpainting.} $\method_{inpainting}$ model can seamlessly paint one's face onto another person under the control of a language prompt. As shown in \cref{fig:inpainting}, by utilizing four reference images of ``Leonardo DiCaprio'' along with language prompts, the model ensures the preservation of facial identity while introducing various facial details. Prompts may encompass modifications in facial expressions, as well as the addition of glasses or a beard.

\vspace{-3mm}
\subsection{Benchmark Results}
\vspace{-1mm}
We create a benchmark with the 200  ID clusters in our validation split. We select 4 reference images and 1 target image from each cluster.  To evaluate the generated images, we measure the face similarity using a face recognition model (ArcFace-R101~\cite{Deng_2022}). The most important metric is the face similarity between the generated image and the target image, denoted as $Sim_{Target}$. Additionally, we calculate the maximum face similarity between the generated images and all reference faces, denoted as $Sim_{Ref}$.  We also report the CLIP similarity score.  To demonstrate that the model can extract ID information and vary following language prompts instead of simply pasting a reference face onto the target image, we calculate $Paste=Sim_{Ref} - Sim_{Target}$ as a metric. A higher value of $Paste$ indicates a greater likelihood of pasting one of the reference images. For the single reference image method, we simply sample the first reference image as input. We also report the FID score~\cite{heusel2018gans, parmar2022aliased} as a metric of image quality.

\begin{table}[t]
    \centering
    \caption{%
    \textbf{Performance comparison of \method and previous methods.}
        We report the face similarity score between the generated image and the target image($Sim_{Target}$), as well as the maximum similarity between the generated face and all reference faces ($Sim_{Ref}$). We also provide the $CLIP$ similarity score of the image. A higher value of $Paste=Sim_{Ref} - Sim_{Target}$ indicates a greater likelihood of pasting one reference face. Our \method achieves the best results in terms of $Sim_{Target}$, demonstrating a high fidelity of ID and better language following ability.
    }
    \label{tab:bench}
    \vspace{-8pt}
    \setlength{\tabcolsep}{6pt}

    \begin{tabular}{lccccc}
        \toprule
        Method & \textit{$Sim_{Ref}$}    & \textit{CLIP $\uparrow$} & \textit{Paste $\downarrow$} & \textit{$Sim_{Target}\uparrow$} & \textit{FID $\downarrow$}  \\
        \midrule    
        IP-Adaptor(SD1.5)~\cite{ye2023ip-adapter} &33.4  & 57.5 & 7.7  & 25.7 & 121.1 \\
        FastComposer(SD1.5)~\cite{xiao2023fastcomposer} &33.3  & 65.2 & 7.5 & 25.8 & 104.9 \\
        PhotoMaker(SDXL)~\cite{li2023photomaker} &45.8  & 72.1 & 5.8 & 40.0 & 84.2 \\
        InstantID(SDXL)~\cite{wang2024instantid} &68.6   & 62.7 & 16.0 & 52.6 & 106.8 \\
        \midrule
        $\method(SD1.5)$& 63.7   & 74.2 & 7.6 & \textbf{56.1} & 85.5\\
        $\method(SD1.5)_{inpaiting}$& 65.6   & 89.2 & 3.8 & \textbf{61.8} & 26.6 \\
        \bottomrule
    \end{tabular}
    \vspace{-10pt}
\end{table}

As shown in \cref{tab:bench},  IP-Adapter~\cite{ye2023ip-adapter}, FastComposer~\cite{xiao2023fastcomposer} and PhotoMaker~\cite{li2023photomaker} demonstrate a low score for both $Sim_{Ref}$ and $Sim_{Target}$, indicating challenges in preserving facial details when using their embedding-based pipeline. For our concurrent work, InstantID~\cite{wang2024instantid}, a very high 
value \textit{Paste} suggests a strong tendency of reference face pasting instead of following the language prompt. Our method outperforms others in terms of $Sim_{Targe}$ with a reasonable \textit{Paste}. This suggests that the \method is capable of doing facial variations based on the language prompt while maintaining a high level of identity fidelity.

\subsection{Ablation study}
\noindent \textbf{Details preserving ability of ReferenceNet.}
We assess the unbounded ability of ReferenceNet to maintain face details. We train our model using the cropped target face as the input reference, similar to IP-Adapter~\cite{ye2023ip-adapter} and FastComposer~\cite{xiao2023fastcomposer}. As shown in ~\cref{tab:reconstruct}, \method can achieve much higher $Sim_{Ref}$ than other methods, indicating the strong ability of ReferenceNet to preserve face details.

\begin{table}[t]
    \centering
    \caption{%
     \textbf{Identity-preserving comparison under the same training paradigm.}  When employing the same training paradigm as previous methods, which involve cropping the human face from the target as a reference, \method achieves a high $Sim_{Ref}$. This indicates \method has a strong ability to preserve details.
    }
    \label{tab:reconstruct}
    \vspace{-8pt}
    \setlength{\tabcolsep}{7.2pt}
    \begin{tabular}{lcccc}
        \toprule
        & IP-Adaptor~\cite{ye2023ip-adapter} & FastComposer~\cite{xiao2023fastcomposer} & Instant-ID~\cite{wang2024instantid} & \method \\
        \midrule
        $Sim_{Ref}$    &  33.4 & 33.3  & 68.6 & 82.0 \\
        \bottomrule
    \end{tabular}
    \vspace{-10pt}
\end{table}

\begin{figure}[!h]
    \centering
        \vspace{-8pt}
    \includegraphics[width=0.8\linewidth]{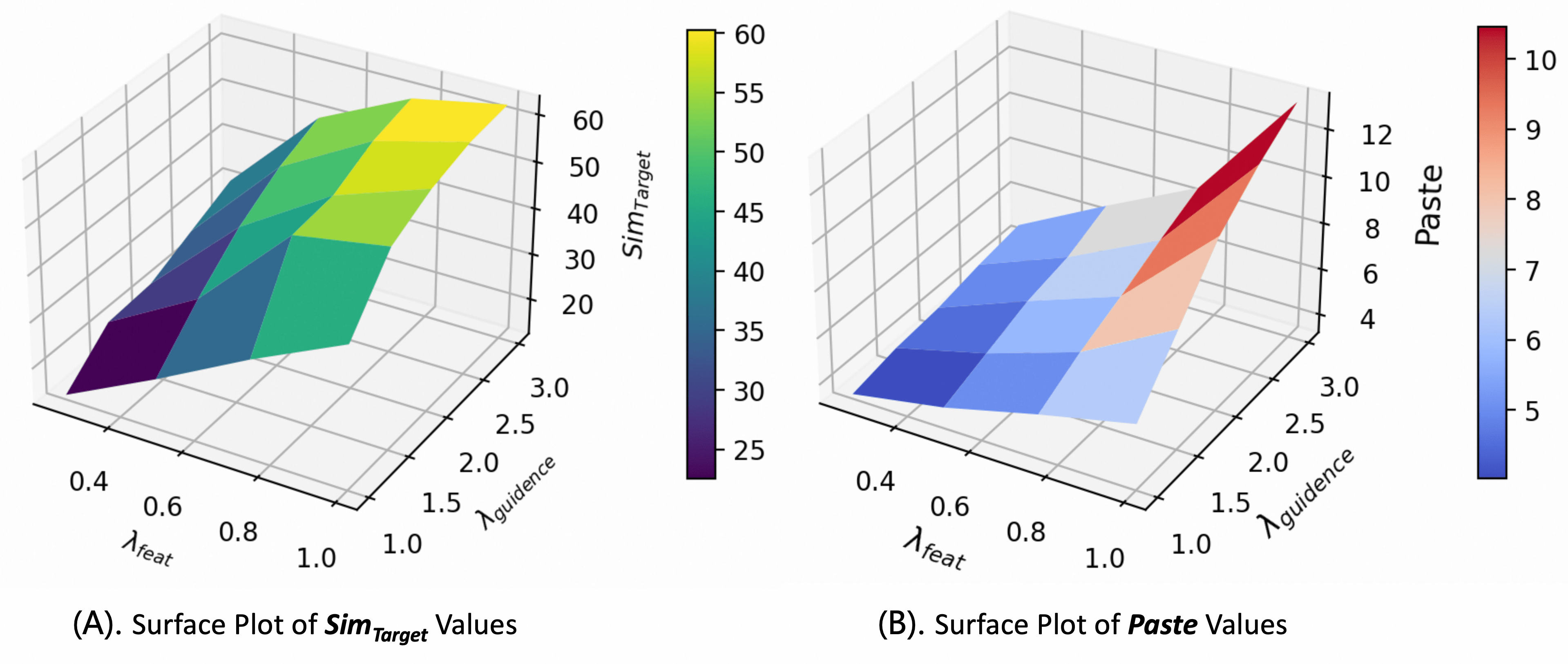}
    \vspace{-10pt}
    \caption{%
    \textbf{The surface plot of $Sim_{Target}$ and $ Paste$ with respect to $\lambda_{feat}$ and $\lambda_{guidance}$.} Both values, $Paste$ and $Sim_{Target}$, increase as $\lambda_{feat}$ and $\lambda_{guidance}$ increase. To balance fidelity and language prompting ability, we recommend setting $\lambda_{feat}$ between 0.75 and 1, and $\lambda_{guidence}$ between 2 and 2.5.
    }
    \label{fig:strength_plot}
    \vspace{-10pt}
\end{figure}

\noindent \textbf{Face reference strength.}  \cref{fig:strength_plot}.(A) reveals a  fidelity($Sim_{Target}$) improvement when enhancing the reference strength with $\lambda_{feat}$ and $\lambda_{guidence}$. However, the $Paste$ also increases as shown in \cref{fig:strength_plot}.(B). This suggests that there is also an increasing tendency to copy the reference image rather than incorporating language prompts. To achieve a balance between fidelity and language prompting ability, we suggest setting $\lambda_{feat}$ between 0.75 and 1, and $\lambda_{guidence}$ between 2 and 2.5.

\label{sec:ref_strength}

\begin{figure}[t]
    \centering
    \includegraphics[width=1\linewidth]{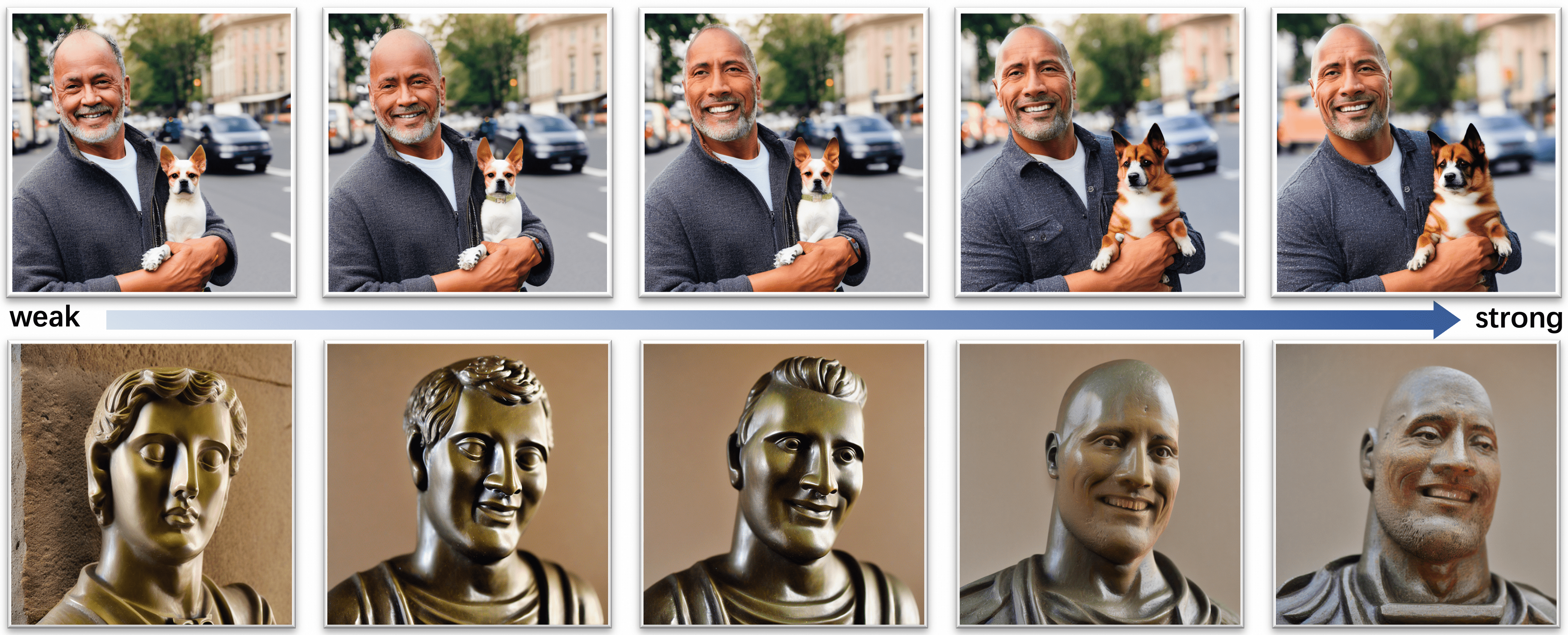}
    \vspace{-10pt}
    \caption{%
     \textbf{Smooth ID fidelity improvement as $\lambda_{feat}$ ranges from 0 to 1.} Using four reference face images of ``Dwayne Johnson'', we increase the  $\lambda_{feat}$  from 0 to 1, moving from left to right, while $\lambda_{guidence}$ is set to 2. Throughout this process, we notice a smooth improvement in ID fidelity.
    }
    \label{fig:ref_strength}
    \vspace{-8pt}
\end{figure}

We also note that $\lambda_{feat}$ exhibits smoother $Sim_{Target}$ changes as it operates in the feature space. We visualize the results for $\lambda_{feat}$ ranging from 0 to 1 under $\lambda_{guidence}=2$ to show the smooth visual similarity improvement (as shown in \cref{fig:ref_strength}). Notably, our model consistently produces natural results across the entire  $\lambda_{feat}$  range. Another noteworthy observation is that in cases of conflicting conditions, such as when attempting to create a bronze-like face of ``Dwayne Johnson'', the optimal result is not achieved at  $\lambda_{feat}=1$. At that point, the facial features lose their metallic luster. This highlights an advantage of our framework: the ability to balance conflicts by adjusting the control strength between the text and reference. In our \supp, we demonstrate the smooth change characteristics of $\lambda_{feat}$ for identity mixing and highlight the role of $\lambda_{guidence}$ in preserving details.

\noindent \textbf{Number of input reference faces.} \cref{tab:num_ref} presents the evaluation results of \method using different numbers of reference images. Increasing the number of reference faces leads to a noticeable improvement in identity enhancement. Specifically, the improvement is significant when changing from 1 reference face to 2, resulting in a 9.3\% increase. Another crucial aspect is the reduction of $Paste$. This reduction implies that the model can learn more precise identity information and generate high-quality images based on the prompt, rather than simply relying on reference images.

\begin{table}[t]
    \centering
    \caption{%
        \textbf{Performance of \method under different numbers of reference images.} Increasing the number of reference images led to a significant improvement in face fidelity.
    }
    \label{tab:num_ref}
    \vspace{-8pt}
    \setlength{\tabcolsep}{12.5pt}
    \begin{tabular}{cccccc}
        \toprule
        \# Ref & \textit{$Sim_{Ref}$}    & \textit{CLIP $\uparrow$} & \textit{Paste $\downarrow$} & \textit{$Sim_{Target}\uparrow$} & \textit{FID $\downarrow$}  \\
        \midrule
        1 & 49.9 & 69.6 & 9.8 & 40.1 &  85.3 \\
        2 & 58.7  & 72.2 & 9.2 & 49.4 & 85.1 \\
        3 & 62.3   & 73.4 & 8.0 & 54.3 & 85.4 \\
        4 & 63.7   & 74.2 & 7.6 & 56.1 & 85.5 \\
        \bottomrule
    \end{tabular}
    \vspace{-10pt}
\end{table}

\begin{figure}[t]
    \centering
    \includegraphics[width=0.8\linewidth]{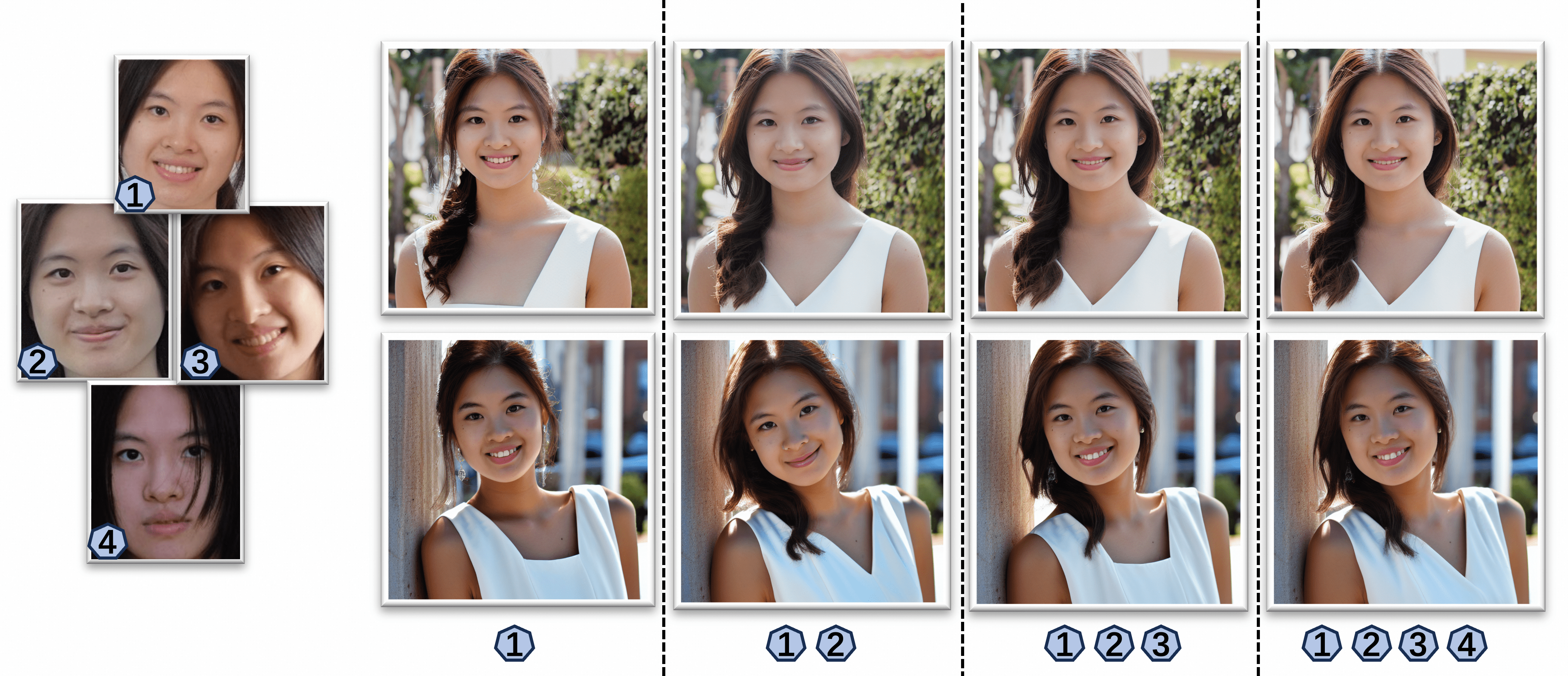}
    \vspace{-12pt}
    \caption{%
       \textbf{Generated image visualization under different numbers of reference faces.} The number of reference images increases from 1 to 4, progressing from left to right. We observe a significant improvement in fidelity, particularly when changing the number from 1 to 2.
    }
    \label{fig:num_ref}
    \vspace{-10pt}
\end{figure}

The consistent visualization results can be found in \cref{fig:num_ref}, which contains generated images with the prompt  ``A young woman in a white dress''. Increasing the number of reference images from 1 to 2 indeed significantly enhances the fidelity of identity. When only one reference image is used, the model can capture certain facial features such as the nose and eyes. However, it struggles to accurately depict the overall face shape(The generated face is not round enough). Nevertheless, as the number of reference images increases, the model gradually improves in capturing all these details.

\noindent \textbf{Position of the reference attention layer.} As shown in \cref{tab:position}, we conduct ablation experiments by adding reference attention layers to different positions of the U-Net, including the encoder blocks, decoder blocks, or both(Adding reference attention layer to middle blocks by default). We observe that when reference attention layers are only added to the encoder blocks, the model is hard to capture the face identity. This can be attributed to the encoder being further away from the output, making it difficult to align with the convergence direction indicated by the loss. Corresponding to this, when reference attention layers are added solely to the decoder blocks, we achieve the best performance even compared to adding them in both the encoder and decoder.

\begin{table}[t]
    \centering
    \caption{%
    \textbf{Performance of incorporating reference attention layers to various positions in the U-Net.}  
  We achieve the best performance by adding it to the decoder blocks.
    }
    \label{tab:position}
    \vspace{-8pt}
    \setlength{\tabcolsep}{8pt}
    \begin{tabular}{lccccc}
        \toprule
        Model Part & \textit{$Sim_{Ref}$}    & \textit{CLIP $\uparrow$} & \textit{Paste $\downarrow$} & \textit{$Sim_{Target}\uparrow$} & \textit{FID $\downarrow$}  \\
        \midrule
        Encoder & 8.4 & 61.9 & 1.6 & 6.8 &  90.3 \\
        Encoder \& Decoder & 60.3  & 73.0 & 8.1 & 52.2 & 88.1 \\
        Decoder & 63.7   & 74.2 & 7.6 & 56.1 & 85.5\\
        \bottomrule
    \end{tabular}
    \vspace{-13pt}
\end{table}

%% file: sections/5.conclusion.tex
\vspace{-5mm}
\section{Conclusion}

In this paper, we present \method, a zero-shot human image personalization method. It excels in \textit{higher-fidelity identity preservation} and \textit{better instruction following}. We encode the reference face into a series of feature maps of different spatial shapes with a Face ReferenceNet, which effectively preserves both face shape and details. Additionally, we utilize a disentangled condition integration strategy that balances the text and image guidance during the text-to-image generation process, thereby achieving powerful language control even if there are conflicts between the reference image and language prompt.  Experiments prove we achieve exceptional personalization capabilities, offering endless possibilities for related applications.

\newpage

%% file: sections/6.ref.tex
\bibliographystyle{splncs04}
\bibliography{ref}

%% file: sections/7.appendix.tex
\begin{itemize}
  \item~\cref{sec:ref_images}: All the reference images for each figure in the main script.
  \item~\cref{sec:statistics}: More statistics about the dataset.
  \item~\cref{sec:id_mix}: Identity mixing visualization.
  \item~\cref{sec:lamda_ref}: How the $\lambda_{ref}$ helps to preserve more details.
  \item~\cref{sec:artwork-real}: Comparison of artwork-real transformation with PhotoMaker.
  \item~\cref{sec:lora}: LoRA version of \method.
   \item~\cref{sec:pos_mask}: Does face position mask affect performance of \method?
  \item~\cref{sec:more_vis}: More visualization results.
  \item~\cref{sec:limitation}: Limitations and Social Impact.
\end{itemize}

In this \supp, we begin by showcasing all the reference images for each figure in the main script. Then, we provide more statistics about the dataset we collected. Additionally, we present identity mixing visualization results to demonstrate the smooth control of $\lambda_{feat}$ in the feature space. Then, we show how the $\lambda_{ref}$ helps to preserve more details. Furthermore, we compare the transformation between artwork and real people with the previous state-of-the-art PhotoMaker~\cite{li2023photomaker}.  We also report the performance without the face position mask. Then we provide a LoRA~\cite{hu2021lora} version of \method and analyze its performance.  We provide more visualization results that demonstrate our strong ability to maintain identity fidelity and follow language prompts across various applications under diverse prompts.
Finally, we discuss our limitations, social impact, and responsibility to human subjects.

\section{Reference Images for Each Figure}
\label{sec:ref_images}
\begin{figure}[!h]
\vspace{-8mm}
\centering
\includegraphics[width=1\linewidth]{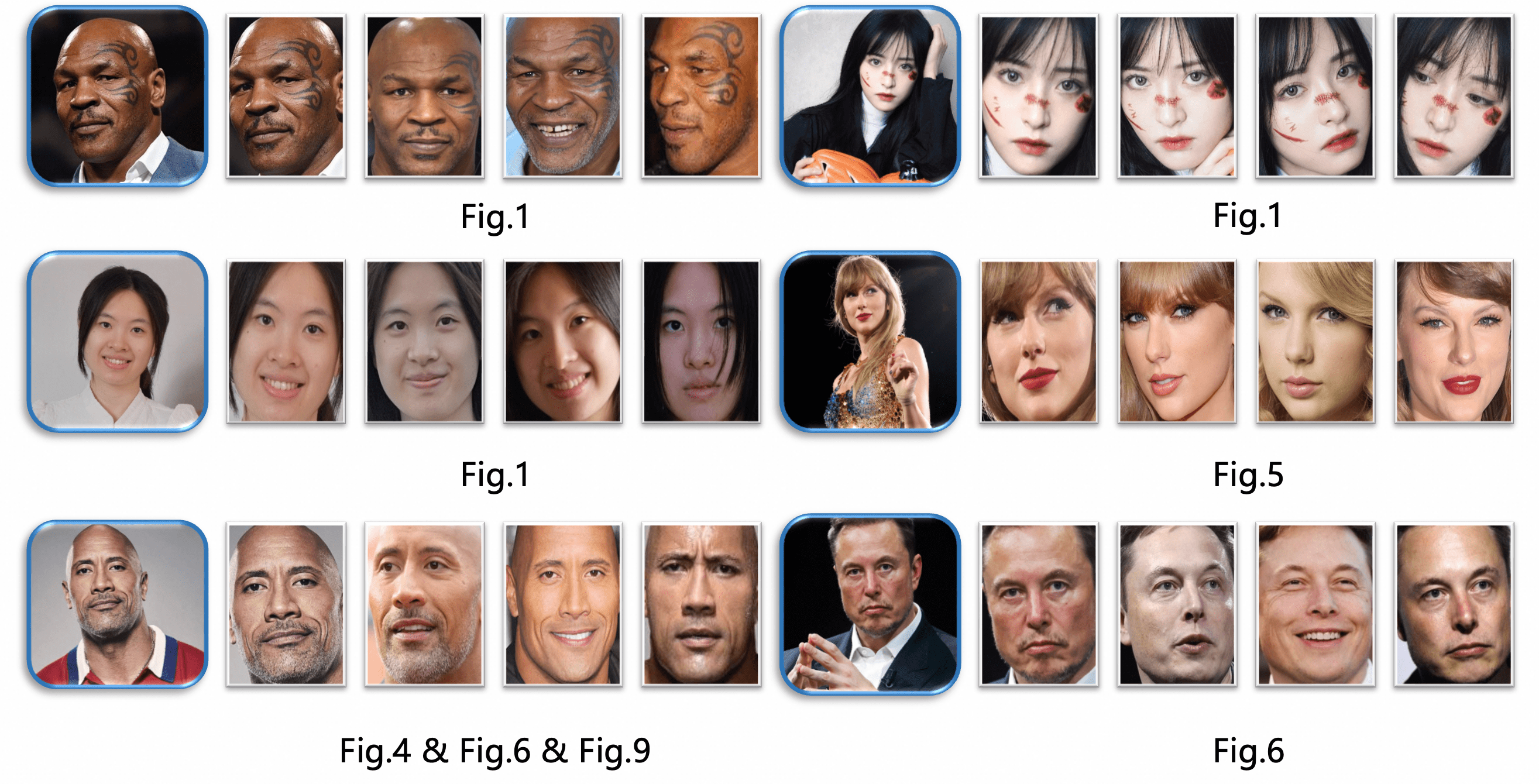}
\caption{
\textbf{Reference faces images for each figure in the main script.} Most generated images have four reference images, while the virtual person to real person transformation has only one reference face image, which is cropped from the ID Images. Therefore, we omit them in this figure.}

\label{fig:supp_input}
\end{figure}
As shown in~\cref{fig:supp_input}, We list all reference images for each individual who is presented in the figures of the main script. Real-person customization involves four reference images, while virtual person to real person transformation only requires one reference face image, cropped from the ID Images. For brevity,  we omit the reference images for the virtual person in Figure~\ref{fig:supp_input}. 

\section{Statistics about the Dataset}
\label{sec:statistics}
\begin{figure}[!h]
\centering
\includegraphics[width=0.7\linewidth]{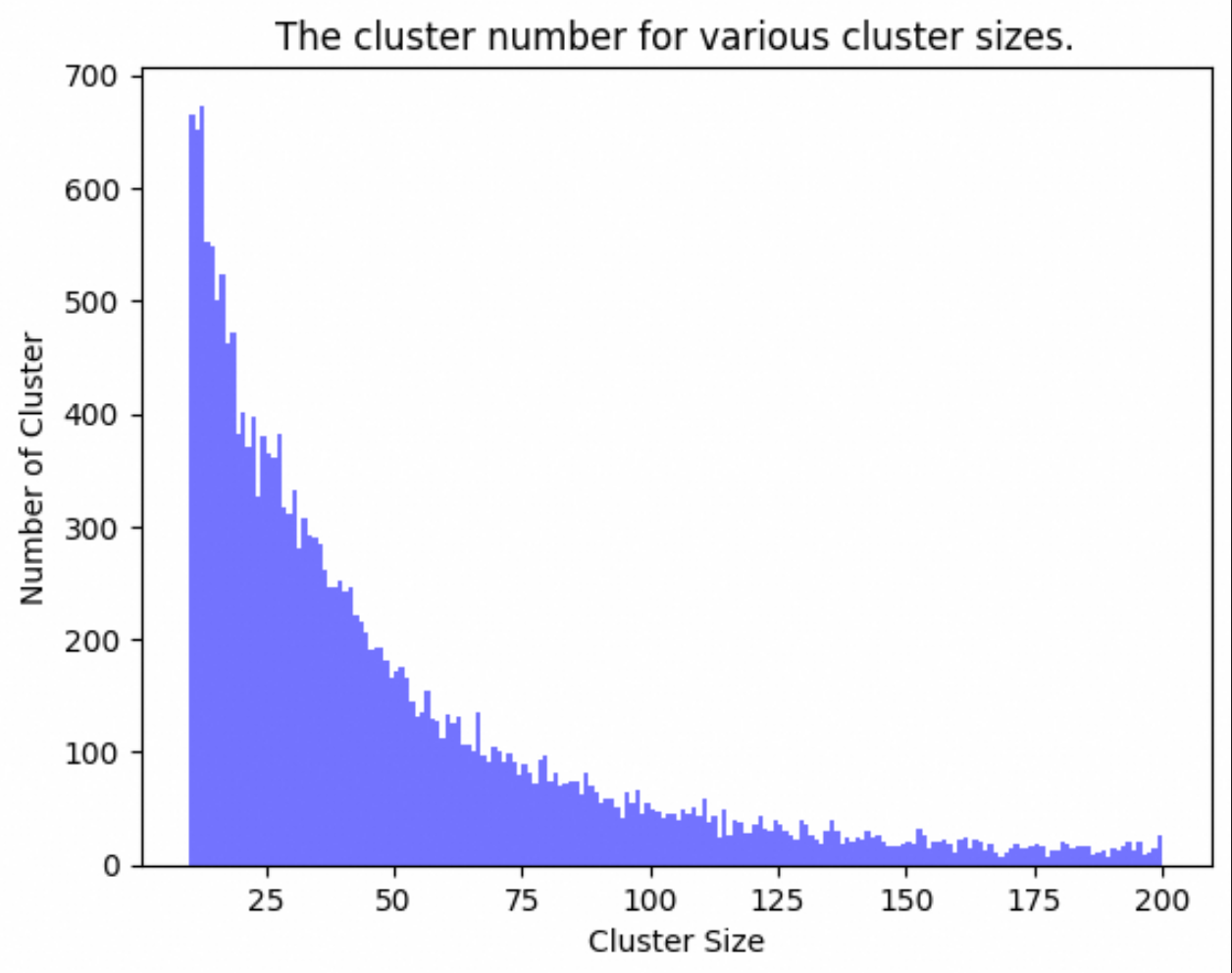}
\caption{
\textbf{Number of clusters for different cluster size} }
\label{fig:stastics}
\end{figure}

\begin{figure}[!h]
\centering
\includegraphics[width=\linewidth]{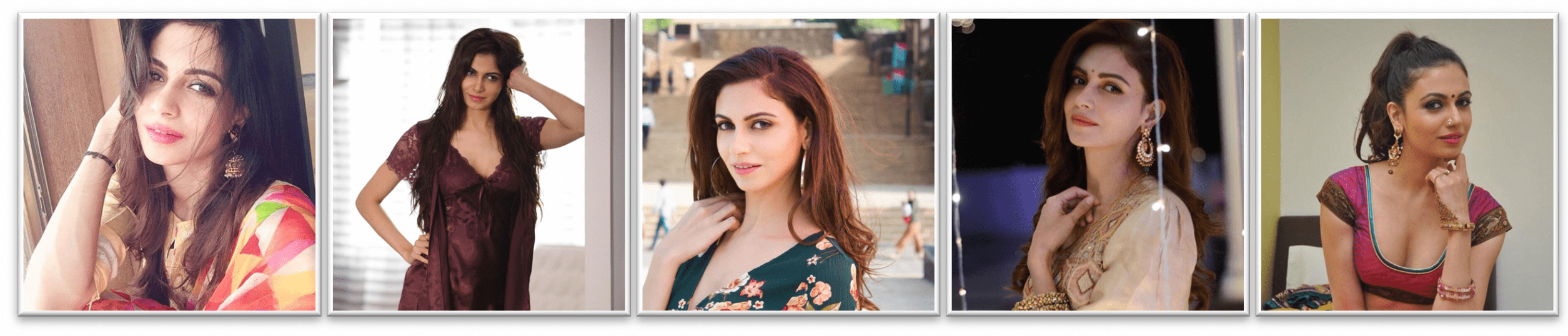}
\caption{
\textbf{Five images from a cluster} }
\label{fig:dataset_example}
\end{figure}

We provide statistics on the dataset collected using our ID data collection pipeline, which consists of 1.8 million images featuring 23,042 individuals. In the~\cref{fig:stastics}, we display the number of individuals with specific cluster sizes. In~\cref{fig:dataset_example}, we also provide five images from the same cluster to demonstrate the variance within the cluster.

\section{Identity Mixing}
\label{sec:id_mix}
We do the identity-mixing by adjusting the weights of the reference layer attention results for two different individuals with $\lambda_{ID1}$ and $\lambda_{ID2}$(as shown in~\cref{equ:id_mix},). These weights are constrained by the equation $\lambda_{ID1} + \lambda_{ID2} = 1$. Additionally, $y^{id_i}_{rj}$ represents the $j$-th reference face feature of the $i$-th person . 

\begin{equation}
 \lambda_{ID1} * Attn(y_{g}, cat[y_{g}, y^{id_1}_{r1}, y^{id_1}_{r2} ...]) + \lambda_{ID2} *Attn(y_{g}, cat[y_{g}, y^{id_2}_{r1}, y^{id_2}_{r2} ...])
\label{equ:id_mix}
\end{equation}

We visualize the results of mixing two individuals, ``Taylor Swift'' and ``Leonardo DiCaprio''. By increasing the weight of ``Leonardo DiCaprio'' from left to right, we observe a smooth transition from ``Taylor Swift'' to ``Leonardo DiCaprio''.

\begin{figure}[!h]
\centering
\includegraphics[width=1\linewidth]{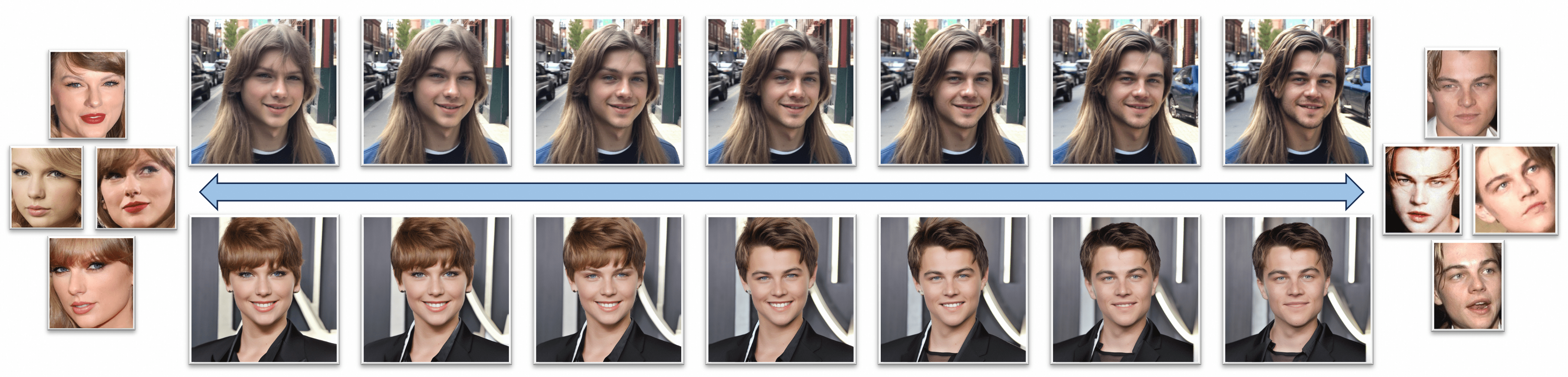}
\caption{
\textbf{Visualization of identity mixing results.} We present the smooth transition from "Taylor Swift" to "Leonardo DiCaprio" in the identity mixing process, displayed from left to right.}
\label{fig:id_mix}
\end{figure}

\section{Visualization under Different  $\lambda_{ref}$ and  $\lambda_{feat}$}
\label{sec:lamda_ref}
\begin{figure}[!h]
\centering
\includegraphics[width=1\linewidth]{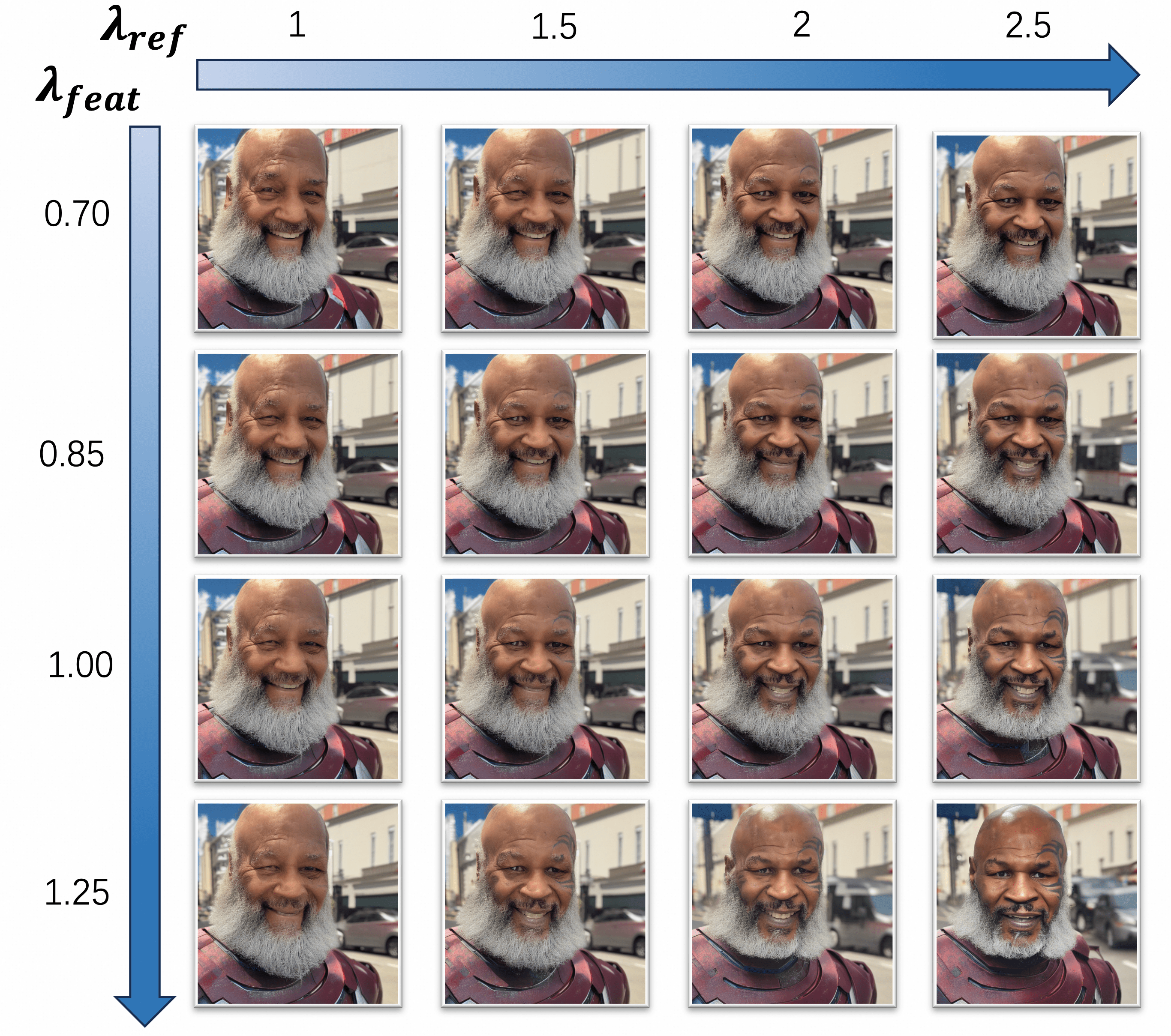}
\caption{
\textbf{Visualization under different values of $\lambda_{ref}$ and $\lambda_{feat}$.}}
\label{fig:strength_vis}
\end{figure}

We present the visualization results of "Mike Tyson" with varying values of $\lambda_{ref}$ and $\lambda_{feat}$ in the~\cref{fig:strength_vis}. Increasing $\lambda_{ref}$ and $\lambda_{feat}$ leads to a noticeable increase in fidelity. 

\begin{table}[!h]
    \centering
    \caption{%
    \textbf{Performance of different versions of \method.} We  report the performance of the LoRA version of \method($\method_{LoRA}(SD1.5)$). Additionally, we evaluate the performance when removing the reference classifier free guidance($\method_{w/o-cfg}(SD1.5)$) and when removing the face position mask($\method_{w/o-pmask}(SD1.5)$). We report the face similarity score between the generated image and the target image($Sim_{Target}$), as well as the maximum similarity between the generated face and all reference faces ($Sim_{Ref}$). We also provide the $CLIP$ similarity score of the image. A higher value of $Paste=Sim_{Ref} - Sim_{Target}$ indicates a greater likelihood of pasting one reference face. 
    }
    \label{tab:more_results}
    \vspace{-8pt}
    \setlength{\tabcolsep}{6pt}
    \begin{tabular}{lccccc}
        \toprule
        Method & \textit{$Sim_{Ref}$}    & \textit{CLIP $\uparrow$} & \textit{Paste $\downarrow$} & \textit{$Sim_{Target}\uparrow$} & \textit{FID $\downarrow$}  \\
        \midrule    
        IP-Adaptor(SD1.5)~\cite{ye2023ip-adapter} &33.4  & 57.5 & 7.7  & 25.7 & 121.1 \\
        FastComposer(SD1.5)~\cite{xiao2023fastcomposer} &33.3  & 65.2 & 7.5 & 25.8 & 104.9 \\
        PhotoMaker(SDXL)~\cite{li2023photomaker} &45.8  & 72.1 & 5.8 & 40.0 & 84.2 \\
        InstantID(SDXL)~\cite{wang2024instantid} &68.6   & 62.7 & 16.0 & 52.6 & 106.8 \\
        \midrule
        $\method(SD1.5)$& 63.7   & 74.2 & 7.6 & \textbf{56.1} & 85.5\\
        $\method_{w/o-pmask}(SD1.5)$& 62.3   & 73.2 &  7.1 & \textbf{55.2} & 85.8\\
        $\method_{w/o-cfg}(SD1.5)$& 61.6   & 72.9 & 8.3 & \textbf{53.3} & 86.1\\
        $\method_{LoRA}(SD1.5)$& 48.7   & 71.7 & 5.3 & \textbf{43.4} & 89.1\\

        \bottomrule
    \end{tabular}
    \vspace{-10pt}
\end{table}

\begin{figure}[!h]
\centering
\includegraphics[width=1\linewidth]{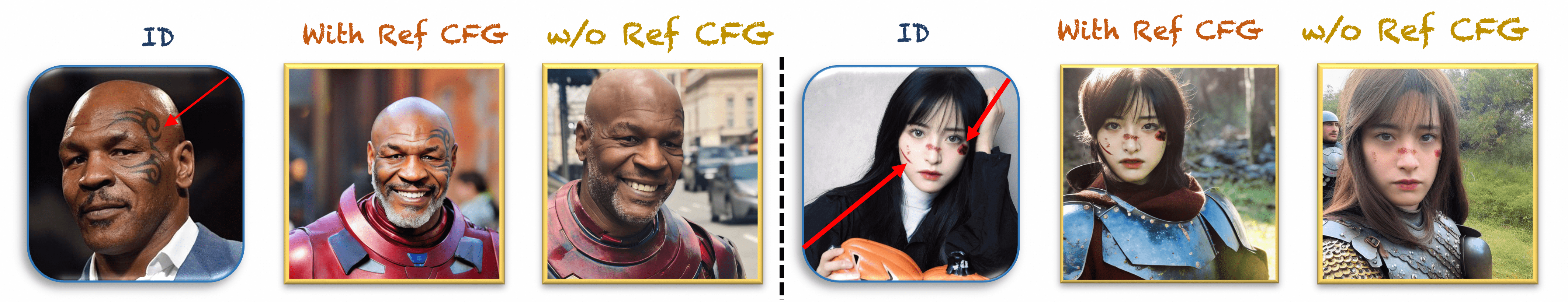}
\caption{
\textbf{Visualization of with and without reference classifier-free guidance.}}
\label{fig:cfg}
\end{figure}

An interesting observation is that $\lambda_{ref}$ is crucial for preserving details. When we remove the reference classifier-free guidance~\cite{ho2022classifier} and use the inference method as ~\cref{equ:ref_guidence}, we notice that the face shape remains faithful to the identity, and get pretty good results for face similarity($\method_{w/o-cfg}(SD1.5)$ in~\cref{tab:more_results}). However, certain details such as scars, tattoos, and other face details are challenging to preserve(as shown in~\cref{fig:cfg}).

\begin{equation}
Y = Y_{Ref} +  \lambda_{text} * (Y_{Text\&Ref} - Y_{Ref}) 
\label{equ:ref_guidence}
\end{equation}

\section{Comparison of  Artwork and Real People Transformation.}
\label{sec:artwork-real}
\begin{figure}[!h]
\centering
\includegraphics[width=1\linewidth]{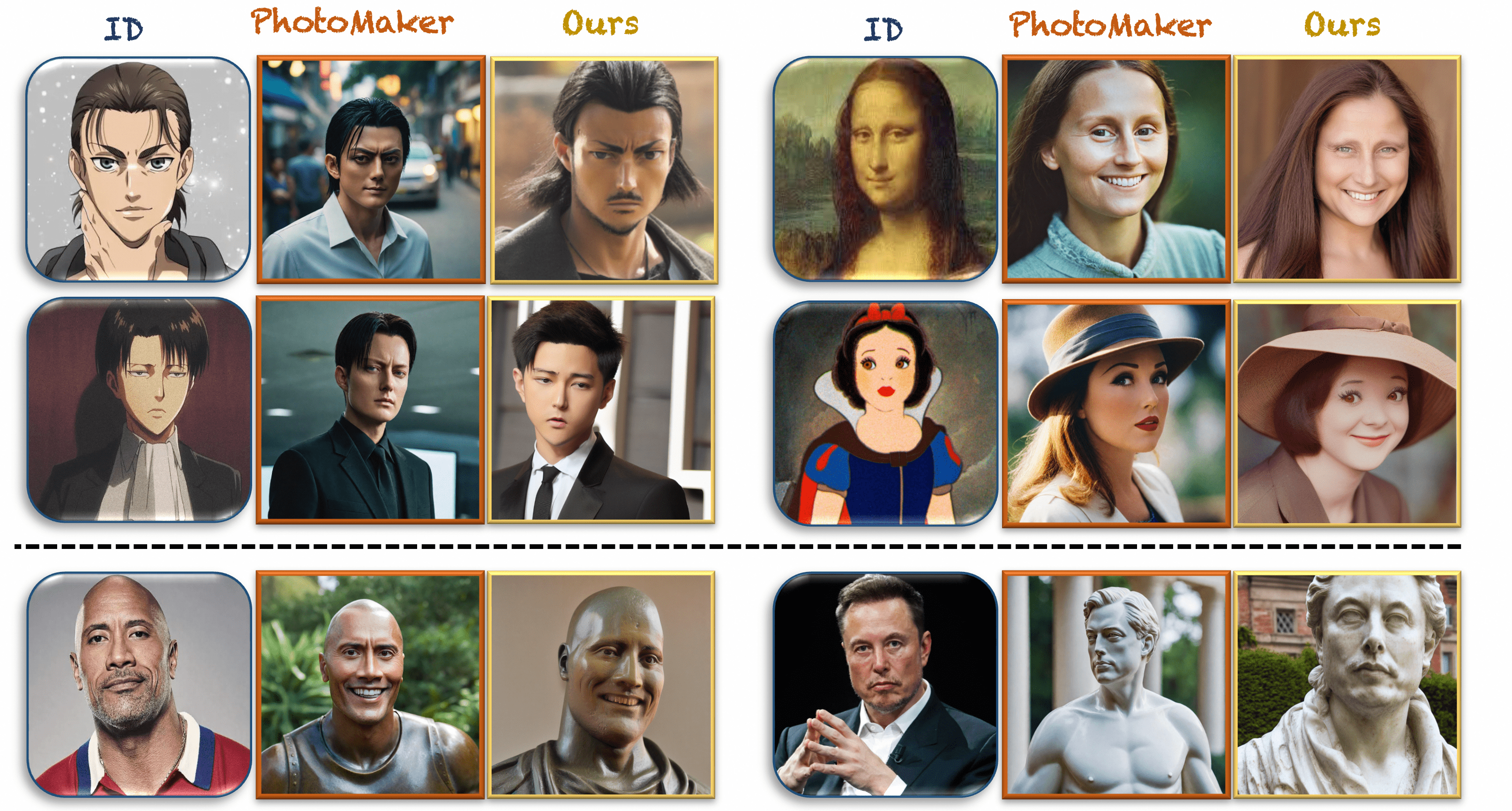}
\caption{
\textbf{Comparison of  artwork and real People transformation.} }
\label{fig:comparison_art}
\end{figure}

We compare our results with the previous state-of-the-art  PhotoMaker~\cite{li2023photomaker}. Using the same input reference image and prompt, we present the results in Figure~\ref{fig:comparison_art}. Our method achieves comparable or even superior results compared to the previous state-of-the-art, utilizing a more compact model, SD-1.5~\cite{rombach2021highresolution}.

\section{LoRA Version of \method}
\label{sec:lora}
We additionally provide a LoRA version of \method by integrating LoRA into all attention layers of the U-Net. We maintain the same hyperparameters except for changing the resolution to 512$*$512. We solely train the LoRA, reference attention layers and face referencenet. As $\method_{LoRA}(SD1.5)$ in~\cref{tab:more_results}, this model also attains competitive results. 

\begin{figure}[!h]
\centering
\includegraphics[width=\linewidth]{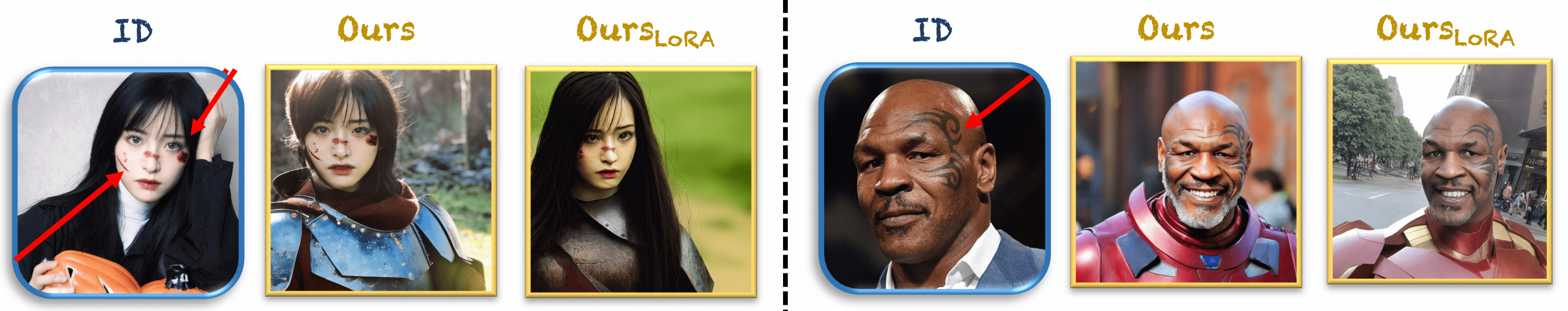}
\caption{
\textbf{Visualization of LoRA version of \method }}
\label{fig:lora_details}
\end{figure}
We provide a visualization of LoRA versions in ~\cref{fig:lora_details}.  Due to its resolution limitation of 512$*$512, we can observe that its ability to preserve face shape is still satisfactory. But its image quality and ability to preserve details are inferior to the full fine-tuning version.

\section{Face Position Mask}
\label{sec:pos_mask}

As $\method_{w/o-pmask}(SD1.5)$ in the ~\cref{tab:more_results}, it can be observed that the face position mask has minimal impact on performance(we fill all zero values in the additional mask channel for this ablation). This aligns with our earlier mention that it is an optional user input primarily used for user-controlled composition.

\section{Additional Visualization Results}
\label{sec:more_vis}
\begin{figure}[!h]
\centering
\includegraphics[width=\linewidth]{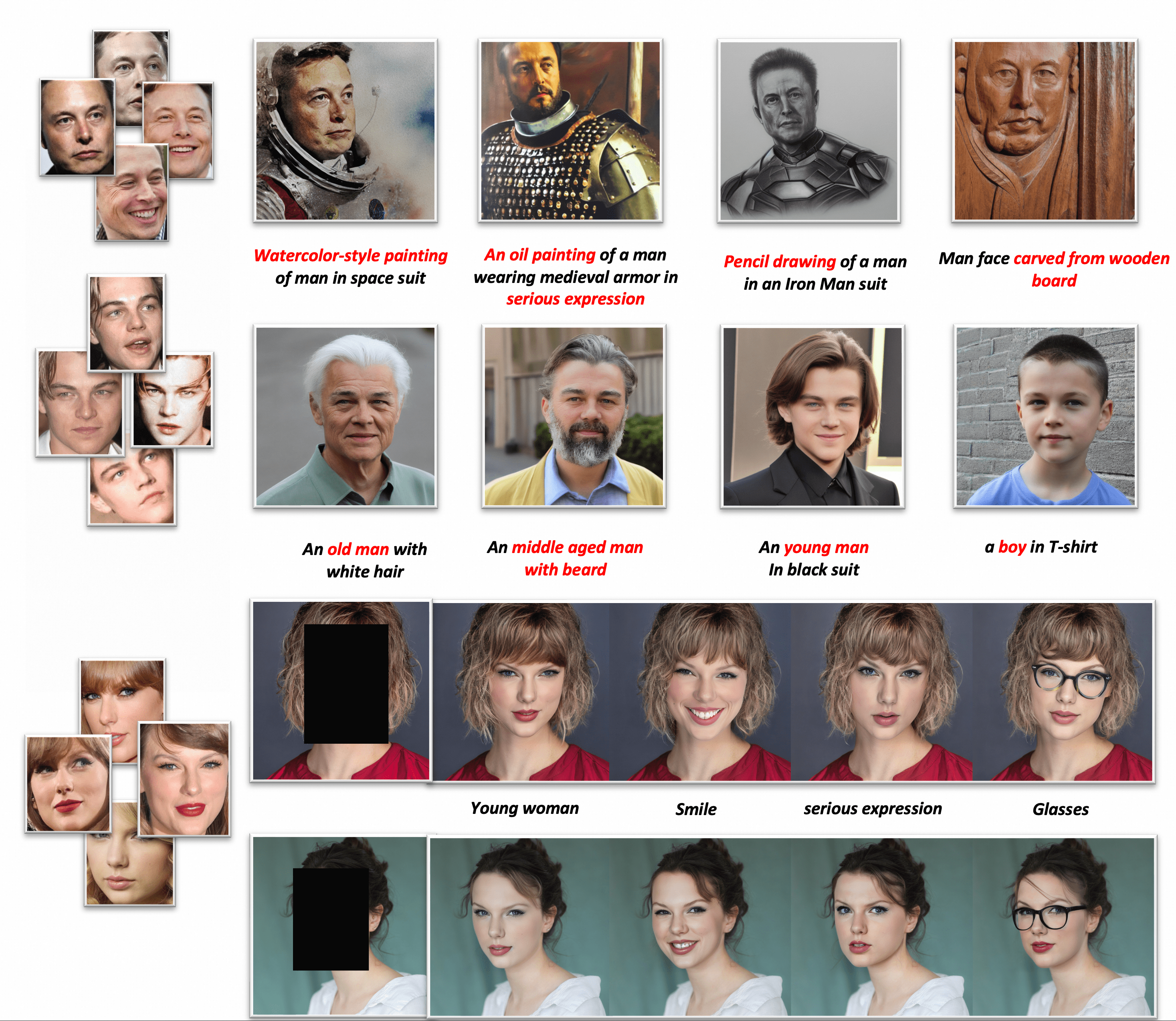}
\caption{
\textbf{More visualization of \method }}
\label{fig:more_vis}
\end{figure}

We present more visualizations of \method in~\cref{fig:more_vis}. These include transforming ``Elon Musk'' into artwork, altering the age of ``Leonardo DiCaprio'', and inpainting ``Taylor Swift'' face with different prompts.

\section{Limitations and Social Impact}
\label{sec:limitation}

\noindent\textbf{Limitations.} Our method has limitations in terms of success rate, as artifacts may appear in some images. This could be attributed to the base model SD-1.5~\cite{rombach2021highresolution}, we may try to change to the bigger model SDXL~\cite{podell2023sdxl}. Furthermore, we find that describing the pose of the head using language is challenging. In future versions, it may be beneficial to introduce improved controllability.

\noindent\textbf{Social Impact.} As a facial customization method for identity preservation, it can generate realistic character photos, providing an entertaining tool for the public. However, if misused, it has the potential to become a generator of false information. In the future, we should also implement checks on prompts and generated photos to minimize negative impacts.

\noindent\textbf{Responsibility to Human Subjects.} The collected human images in this study are sourced from publicly available images with open copyrights. Since most of them feature public figures, their portraits are relatively less sensitive. Additionally, our data algorithm is solely used for academic purposes and not for commercial use.

\newpage

%% file: ref.bib
@String(CVPR   = {IEEE Conf. Comput. Vis. Pattern Recog.})

@String(ICML   = {Int. Conf. Mach. Learn.})

@misc{radford2016unsupervised,
      title={Unsupervised Representation Learning with Deep Convolutional Generative Adversarial Networks}, 
      author={Alec Radford and Luke Metz and Soumith Chintala},
      year={2016},
      eprint={1511.06434},
      archivePrefix={arXiv},
      primaryClass={cs.LG}
}

@misc{goodfellow2014generative,
      title={Generative Adversarial Networks}, 
      author={Ian J. Goodfellow and Jean Pouget-Abadie and Mehdi Mirza and Bing Xu and David Warde-Farley and Sherjil Ozair and Aaron Courville and Yoshua Bengio},
      year={2014},
      eprint={1406.2661},
      archivePrefix={arXiv},
      primaryClass={stat.ML}
}

@misc{ho2020denoising,
      title={Denoising Diffusion Probabilistic Models}, 
      author={Jonathan Ho and Ajay Jain and Pieter Abbeel},
      year={2020},
      eprint={2006.11239},
      archivePrefix={arXiv},
      primaryClass={cs.LG}
}

@misc{schuhmann2022laion5b,
      title={LAION-5B: An open large-scale dataset for training next generation image-text models}, 
      author={Christoph Schuhmann and Romain Beaumont and Richard Vencu and Cade Gordon and Ross Wightman and Mehdi Cherti and Theo Coombes and Aarush Katta and Clayton Mullis and Mitchell Wortsman and Patrick Schramowski and Srivatsa Kundurthy and Katherine Crowson and Ludwig Schmidt and Robert Kaczmarczyk and Jenia Jitsev},
      year={2022},
      eprint={2210.08402},
      archivePrefix={arXiv},
      primaryClass={cs.CV}
}

@misc{radford2021learning,
      title={Learning Transferable Visual Models From Natural Language Supervision}, 
      author={Alec Radford and Jong Wook Kim and Chris Hallacy and Aditya Ramesh and Gabriel Goh and Sandhini Agarwal and Girish Sastry and Amanda Askell and Pamela Mishkin and Jack Clark and Gretchen Krueger and Ilya Sutskever},
      year={2021},
      eprint={2103.00020},
      archivePrefix={arXiv},
      primaryClass={cs.CV}
}

@inproceedings{ruiz2023dreambooth,
  title={Dreambooth: Fine tuning text-to-image diffusion models for subject-driven generation},
  author={Ruiz, Nataniel and Li, Yuanzhen and Jampani, Varun and Pritch, Yael and Rubinstein, Michael and Aberman, Kfir},
  booktitle={Proceedings of the IEEE/CVF Conference on Computer Vision and Pattern Recognition},
  pages={22500--22510},
  year={2023}
}

@article{gal2022image,
  title={An image is worth one word: Personalizing text-to-image generation using textual inversion},
  author={Gal, Rinon and Alaluf, Yuval and Atzmon, Yuval and Patashnik, Or and Bermano, Amit H and Chechik, Gal and Cohen-Or, Daniel},
  journal={arXiv preprint arXiv:2208.01618},
  year={2022}
}

@article{xiao2023fastcomposer,
  title={FastComposer: Tuning-Free Multi-Subject Image Generation with Localized Attention},
  author={Xiao, Guangxuan and Yin, Tianwei and Freeman, William T and Durand, Fr{\'e}do and Han, Song},
  journal={arXiv preprint arXiv:2305.10431},
  year={2023}
}

@article{ye2023ip-adapter,
  title={IP-Adapter: Text Compatible Image Prompt Adapter for Text-to-Image Diffusion Models},
  author={Ye, Hu and Zhang, Jun and Liu, Sibo and Han, Xiao and Yang, Wei},
  journal={arXiv preprint arxiv:2308.06721},
  year={2023}
}

@misc{Reference-only,
  title={Reference-only controlnet},
  author={Lvmin Zhang},
  howpublished={\url{https://github.com/Mikubill/sd-webui-controlnet/discussions/1236}},
  year={2023.5}
}

@misc{IMDb,
  title={IMDb},
  author={IMDb},
  howpublished={\url{https://www.imdb.com/}},
  year={}
}

@misc{douban,
  title={douban},
  author={douban},
  howpublished={\url{https://m.douban.com/movie/}},
  year={}
}

@misc{YouGov,
  title={YouGov},
  author={YouGov},
  howpublished={\url{https://today.yougov.com/ratings/international/fame/all-time-people/all}},
  year={}
}

@misc{karras2019stylebased,
      title={A Style-Based Generator Architecture for Generative Adversarial Networks}, 
      author={Tero Karras and Samuli Laine and Timo Aila},
      year={2019},
      eprint={1812.04948},
      archivePrefix={arXiv},
      primaryClass={cs.NE}
}

@article{liu2023hyperhuman,
    title={HyperHuman: Hyper-Realistic Human Generation with Latent Structural Diffusion},
    author={Liu, Xian and Ren, Jian and Siarohin, Aliaksandr and Skorokhodov, Ivan and Li, Yanyu and Lin, Dahua and Liu, Xihui and Liu, Ziwei and Tulyakov, Sergey},
    journal={arXiv preprint arXiv:2310.08579},
    year={2023}
}

@article{zheng2021farl,
  title={General Facial Representation Learning in a Visual-Linguistic Manner},
  author={Zheng, Yinglin and Yang, Hao and Zhang, Ting and Bao, Jianmin and Chen, Dongdong and Huang, Yangyu and Yuan, Lu and Chen, Dong and Zeng, Ming and Wen, Fang},
  journal={arXiv preprint arXiv:2112.03109},
  year={2021}
}

@InProceedings{Deng_2020_CVPR,
author = {Deng, Jiankang and Guo, Jia and Ververas, Evangelos and Kotsia, Irene and Zafeiriou, Stefanos},
title = {RetinaFace: Single-Shot Multi-Level Face Localisation in the Wild},
booktitle = {Proceedings of the IEEE/CVF Conference on Computer Vision and Pattern Recognition (CVPR)},
month = {June},
year = {2020}
}

@article{Deng_2022,
   title={ArcFace: Additive Angular Margin Loss for Deep Face Recognition},
   volume={44},
   ISSN={1939-3539},
   url={http://dx.doi.org/10.1109/TPAMI.2021.3087709},
   DOI={10.1109/tpami.2021.3087709},
   number={10},
   journal={IEEE Transactions on Pattern Analysis and Machine Intelligence},
   publisher={Institute of Electrical and Electronics Engineers (IEEE)},
   author={Deng, Jiankang and Guo, Jia and Yang, Jing and Xue, Niannan and Kotsia, Irene and Zafeiriou, Stefanos},
   year={2022},
   month=oct, pages={5962–5979} }

@article{Qwen-VL,
  title={Qwen-VL: A Versatile Vision-Language Model for Understanding, Localization, Text Reading, and Beyond},
  author={Bai, Jinze and Bai, Shuai and Yang, Shusheng and Wang, Shijie and Tan, Sinan and Wang, Peng and Lin, Junyang and Zhou, Chang and Zhou, Jingren},
  journal={arXiv preprint arXiv:2308.12966},
  year={2023}
}

@InProceedings{Rombach_2022_CVPR,
    author    = {Rombach, Robin and Blattmann, Andreas and Lorenz, Dominik and Esser, Patrick and Ommer, Bj\"orn},
    title     = {High-Resolution Image Synthesis With Latent Diffusion Models},
    booktitle = {Proceedings of the IEEE/CVF Conference on Computer Vision and Pattern Recognition (CVPR)},
    month     = {June},
    year      = {2022},
    pages     = {10684-10695}
}

@article{podell2023sdxl,
  title={Sdxl: Improving latent diffusion models for high-resolution image synthesis},
  author={Podell, Dustin and English, Zion and Lacey, Kyle and Blattmann, Andreas and Dockhorn, Tim and M{\"u}ller, Jonas and Penna, Joe and Rombach, Robin},
  journal={arXiv preprint arXiv:2307.01952},
  year={2023}
}

@article{li2023photomaker,
  title={Photomaker: Customizing realistic human photos via stacked id embedding},
  author={Li, Zhen and Cao, Mingdeng and Wang, Xintao and Qi, Zhongang and Cheng, Ming-Ming and Shan, Ying},
  journal={arXiv preprint arXiv:2312.04461},
  year={2023}
}

@article{wang2024instantid,
  title={Instantid: Zero-shot identity-preserving generation in seconds},
  author={Wang, Qixun and Bai, Xu and Wang, Haofan and Qin, Zekui and Chen, Anthony},
  journal={arXiv preprint arXiv:2401.07519},
  year={2024}
}

@misc{kumari2023multiconcept,
      title={Multi-Concept Customization of Text-to-Image Diffusion}, 
      author={Nupur Kumari and Bingliang Zhang and Richard Zhang and Eli Shechtman and Jun-Yan Zhu},
      year={2023},
      eprint={2212.04488},
      archivePrefix={arXiv},
      primaryClass={cs.CV}
}

@article{liu2023cones2,
  title={Cones 2: Customizable Image Synthesis with Multiple Subjects},
  author={Liu, Zhiheng and Zhang, Yifei and Shen, Yujun and Zheng, Kecheng and Zhu, Kai and Feng, Ruili and Liu, Yu and Zhao, Deli and Zhou, Jingren and Cao, Yang},
  journal={arXiv preprint arXiv:2305.19327},
  year={2023}
}

@article{liu2023cones,
  title={Cones: Concept neurons in diffusion models for customized generation},
  author={Liu, Zhiheng and Feng, Ruili and Zhu, Kai and Zhang, Yifei and Zheng, Kecheng and Liu, Yu and Zhao, Deli and Zhou, Jingren and Cao, Yang},
  journal={arXiv preprint arXiv:2303.05125},
  year={2023}
}

@article{chen2023anydoor,
  title={Anydoor: Zero-shot object-level image customization},
  author={Chen, Xi and Huang, Lianghua and Liu, Yu and Shen, Yujun and Zhao, Deli and Zhao, Hengshuang},
  journal={arXiv preprint arXiv:2307.09481},
  year={2023}
}

@article{pan2023kosmos,
  title={Kosmos-g: Generating images in context with multimodal large language models},
  author={Pan, Xichen and Dong, Li and Huang, Shaohan and Peng, Zhiliang and Chen, Wenhu and Wei, Furu},
  journal={arXiv preprint arXiv:2310.02992},
  year={2023}
}

@article{Emu2,
        title={Generative Multimodal Models are In-Context Learners}, 
        author={Quan Sun and Yufeng Cui and Xiaosong Zhang and Fan Zhang and Qiying Yu and Zhengxiong Luo and Yueze Wang and Yongming Rao and Jingjing Liu and Tiejun Huang and Xinlong Wang},
        journal={arXiv preprint arXiv:2312.13286},
        year={2023}
}

@article{wei2023elite,
  title={Elite: Encoding visual concepts into textual embeddings for customized text-to-image generation},
  author={Wei, Yuxiang and Zhang, Yabo and Ji, Zhilong and Bai, Jinfeng and Zhang, Lei and Zuo, Wangmeng},
  journal={arXiv preprint arXiv:2302.13848},
  year={2023}
}

@article{hu2021lora,
  title={Lora: Low-rank adaptation of large language models},
  author={Hu, Edward J and Shen, Yelong and Wallis, Phillip and Allen-Zhu, Zeyuan and Li, Yuanzhi and Wang, Shean and Wang, Lu and Chen, Weizhu},
  journal={arXiv preprint arXiv:2106.09685},
  year={2021}
}

@article{kingma2013auto,
  title={Auto-encoding variational bayes},
  author={Kingma, Diederik P and Welling, Max},
  journal={arXiv preprint arXiv:1312.6114},
  year={2013}
}

@inproceedings{ronneberger2015u,
  title={U-net: Convolutional networks for biomedical image segmentation},
  author={Ronneberger, Olaf and Fischer, Philipp and Brox, Thomas},
  booktitle={Medical Image Computing and Computer-Assisted Intervention--MICCAI 2015: 18th International Conference, Munich, Germany, October 5-9, 2015, Proceedings, Part III 18},
  pages={234--241},
  year={2015},
  organization={Springer}
}

@article{kingma2014adam,
  title={Adam: A method for stochastic optimization},
  author={Kingma, Diederik P and Ba, Jimmy},
  journal={arXiv preprint arXiv:1412.6980},
  year={2014}
}

@article{song2020denoising,
  title={Denoising diffusion implicit models},
  author={Song, Jiaming and Meng, Chenlin and Ermon, Stefano},
  journal={arXiv preprint arXiv:2010.02502},
  year={2020}
}

@article{ho2022classifier,
  title={Classifier-free diffusion guidance},
  author={Ho, Jonathan and Salimans, Tim},
  journal={arXiv preprint arXiv:2207.12598},
  year={2022}
}

@article{yan2023facestudio,
  title={FaceStudio: Put Your Face Everywhere in Seconds},
  author={Yan, Yuxuan and Zhang, Chi and Wang, Rui and Zhou, Yichao and Zhang, Gege and Cheng, Pei and Yu, Gang and Fu, Bin},
  journal={arXiv preprint arXiv:2312.02663},
  year={2023}
}

@misc{heusel2018gans,
      title={GANs Trained by a Two Time-Scale Update Rule Converge to a Local Nash Equilibrium}, 
      author={Martin Heusel and Hubert Ramsauer and Thomas Unterthiner and Bernhard Nessler and Sepp Hochreiter},
      year={2018},
      eprint={1706.08500},
      archivePrefix={arXiv},
      primaryClass={cs.LG}
}

@misc{parmar2022aliased,
      title={On Aliased Resizing and Surprising Subtleties in GAN Evaluation}, 
      author={Gaurav Parmar and Richard Zhang and Jun-Yan Zhu},
      year={2022},
      eprint={2104.11222},
      archivePrefix={arXiv},
      primaryClass={cs.CV}
}

@article{unclip,
  title={Hierarchical text-conditional image generation with clip latents},
  author={Ramesh, Aditya and Dhariwal, Prafulla and Nichol, Alex and Chu, Casey and Chen, Mark},
  journal={arXiv preprint arXiv:2204.06125},
  year={2022}
}

@inproceedings{rombach2021highresolution,
  title={High-resolution image synthesis with latent diffusion models},
  author={Rombach, Robin and Blattmann, Andreas and Lorenz, Dominik and Esser, Patrick and Ommer, Bj{\"o}rn},
  booktitle={IEEE Conference on Computer Vision and Pattern Recognition (CVPR)},
  pages={10684--10695},
  year={2022}
}

@inproceedings{ramesh2021zero,
  title={Zero-shot text-to-image generation},
  author={Ramesh, Aditya and Pavlov, Mikhail and Goh, Gabriel and Gray, Scott and Voss, Chelsea and Radford, Alec and Chen, Mark and Sutskever, Ilya},
  booktitle={Proceedings of the International Conference on Machine Learning (ICML)},
  pages={8821--8831},
  year={2021},
  organization={PMLR}
}

@article{imagen,
  title={Photorealistic text-to-image diffusion models with deep language understanding},
  author={Saharia, Chitwan and Chan, William and Saxena, Saurabh and Li, Lala and Whang, Jay and Denton, Emily and Ghasemipour, Seyed Kamyar Seyed and Ayan, Burcu Karagol and Mahdavi, S Sara and Lopes, Rapha Gontijo and others},
  journal={arXiv preprint arXiv:2205.11487},
  year={2022}
}

@inproceedings{ncsn,
  author       = {Yang Song and
                  Stefano Ermon},
  title        = {Generative Modeling by Estimating Gradients of the Data Distribution},
  booktitle    = {Advances in Neural Information Processing Systems (NeurIPS)},
  pages        = {11895--11907},
  year         = {2019},
}

@inproceedings{sohl2015diffusion,
author       = {Jascha Sohl{-}Dickstein and
                Eric A. Weiss and
                Niru Maheswaranathan and
                Surya Ganguli},
title        = {Deep Unsupervised Learning using Nonequilibrium Thermodynamics},
booktitle    = {Proceedings of the International Conference on Machine Learning (ICML)},
volume       = {37},
pages        = {2256--2265},
year         = {2015},
}

@inproceedings{muse,
author       = {Huiwen Chang and
                Han Zhang and
                Jarred Barber and
                Aaron Maschinot and
                Jos{\'{e}} Lezama and
                Lu Jiang and
                Ming{-}Hsuan Yang and
                Kevin Patrick Murphy and
                William T. Freeman and
                Michael Rubinstein and
                Yuanzhen Li and
                Dilip Krishnan},
title        = {Muse: Text-To-Image Generation via Masked Generative Transformers},
booktitle    = {Proceedings of the International Conference on Machine Learning (ICML)},
pages        = {4055--4075},
year         = {2023},
}

@inproceedings{stylegan,
author       = {Axel Sauer and
                Tero Karras and
                Samuli Laine and
                Andreas Geiger and
                Timo Aila},
title        = {StyleGAN-T: Unlocking the Power of GANs for Fast Large-Scale Text-to-Image Synthesis},
booktitle    = {Proceedings of the International Conference on Machine Learning (ICML)},
pages        = {30105--30118},
year         = {2023},
}

@inproceedings{scalinggan,
author       = {Minguk Kang and
                Jun{-}Yan Zhu and
                Richard Zhang and
                Jaesik Park and
                Eli Shechtman and
                Sylvain Paris and
                Taesung Park},
title        = {Scaling up GANs for Text-to-Image Synthesis},
booktitle    = {IEEE Conference on Computer Vision and Pattern Recognition (CVPR)},
pages        = {10124--10134},
year         = {2023},
}

@inproceedings{dhariwal2021diffusion,
author       = {Prafulla Dhariwal and
                Alexander Quinn Nichol},
editor       = {Marc'Aurelio Ranzato and
                Alina Beygelzimer and
                Yann N. Dauphin and
                Percy Liang and
                Jennifer Wortman Vaughan},
title        = {Diffusion Models Beat GANs on Image Synthesis},
booktitle    = {Advances in Neural Information Processing Systems (NeurIPS)},
pages        = {8780--8794},
year         = {2021},
}

@misc{chen2023gentron,
      title={GenTron: Delving Deep into Diffusion Transformers for Image and Video Generation}, 
      author={Shoufa Chen and Mengmeng Xu and Jiawei Ren and Yuren Cong and Sen He and Yanping Xie and Animesh Sinha and Ping Luo and Tao Xiang and Juan-Manuel Perez-Rua},
      year={2023},
      eprint={2312.04557},
      archivePrefix={arXiv},
      primaryClass={cs.CV}
}
